\ifdefined\pdfminorversion
\fi
\documentclass[journal]{IEEEtran}
\usepackage[T1]{fontenc}
\usepackage[utf8]{inputenc}
\ifdefined\XeTeXversion
  \usepackage{newunicodechar}
  \newunicodechar{–}{\textendash}
  \newunicodechar{—}{\textemdash}
  \newunicodechar{“}{\textquotedblleft}
  \newunicodechar{”}{\textquotedblright}
  \newunicodechar{’}{\textquoteright}
\fi
\usepackage{amsmath,amssymb,bm}
\usepackage{graphicx,booktabs,array,tabularx,longtable}
\usepackage{cite}
\usepackage{dblfloatfix}
\usepackage{placeins}
\usepackage[hidelinks]{hyperref}
\usepackage{url}
\newcommand{\strong}[1]{{\bfseries\boldmath #1}}
\newcommand{\TCell}[4]{#4}
\newcommand{\TNote}[3]{#3}
\newlength{\TableUsable}
\newcommand{\ManuscriptFigure}[1]{%
  \IfFileExists{figures/fig_#1.pdf}{%
    \includegraphics[width=\textwidth]{figures/fig_#1.pdf}%
  }{%
    \fbox{\parbox[c][20mm][c]{0.95\textwidth}{\centering
      Figure #1 placeholder. Upload \texttt{figures/fig\_#1.pdf}.}}%
  }%
}
\begin{document}
\ifdefined\XeTeXversion\fi
\title{QA-CMIF: Accurate and Lightweight Cross-Modal Latent Imputation and Fusion for Fault-Tolerant sEMG–Inertial Hand Gesture Recognition}
\author{Xinyu~Tao, Shuang~Huang\textsuperscript{*}, Yeyun~Cai, Fang~Deng%
\thanks{\textsuperscript{*}Corresponding author: Shuang Huang.}%
\thanks{Xinyu Tao and Shuang Huang are with the Faculty of Marine Science and Technology, Beijing Institute of Technology, Zhuhai 519088, China (e-mail: \href{mailto:huangshuang@bit.edu.cn}{huangshuang@bit.edu.cn}).}%
\thanks{Yeyun Cai is with the National Key Lab of Autonomous Intelligent Unmanned Systems, Beijing Institute of Technology, Beijing 100081, China (e-mail: \href{mailto:caiyeyun@bit.edu.cn}{caiyeyun@bit.edu.cn}).}%
\thanks{Fang Deng is with the School of Artificial Intelligence, Beijing Institute of Technology, Beijing 100081, China (e-mail: \href{mailto:dengfang@bit.edu.cn}{dengfang@bit.edu.cn}).}%
\thanks{This work has been submitted to the IEEE for possible publication. Copyright may be transferred without notice, after which this version may no longer be accessible.}%
}

\maketitle
\bstctlcite{IEEEcontrol}
\begin{abstract}
Fusing surface electromyography (sEMG) with inertial measurements is widely used for myoelectric prosthesis control and wearable human–machine interfaces. In daily wear, however, socket micro-displacement, electrode lift-off, and wire failures often corrupt an unknown subset of sEMG channels through burst artifacts, lift-off shocks, or open-circuit dropouts. Existing fusion networks mix all channels in the first convolutional layer, so a few corrupted electrodes contaminate the entire latent representation, whereas gate-only attenuation of corrupted channels discards the co-activation cues needed to separate fine-grained gestures. We propose QA-CMIF, a quality-aware cross-modal latent imputation and fusion framework that encodes each sEMG channel independently, scores its reliability with a soft gate, and protects intact channels via a confidence-weighted residual bypass, while a cross-attention imputer repairs degraded channels in latent space using the inertial stream and remaining reliable sEMG channels under a bad-channel alignment loss. Across within-subject evaluations on 29 participants from Ninapro DB3, DB2, and DB5 spanning 17--52 gesture classes, QA-CMIF outperformed the gate-only baseline under 50\% channel corruption by 1.29--2.65 percentage points, denoted pp, across all three 17-class cohorts with paired-$t$ Holm-adjusted $p < 0.002$, widened this margin to 2.59--4.15 pp on the 40- and 52-class tasks, and surpassed corruption-augmented conventional fusion by 4.61--19.85 pp. Requiring only $0.175\text{--}0.185\text{M}$ parameters and $2.45\text{--}2.63\text{M}$ multiply--accumulate operations across the $2000\text{ Hz}$ and $200\text{ Hz}$ setups, QA-CMIF saves $46\%\text{--}55\%$ in parameters and $72\%\text{--}81\%$ in compute over recent baselines with a $1.44\text{ ms}$ 99th-percentile single-thread CPU latency.
\end{abstract}

\section{Introduction}

Decoding dexterous hand and wrist motor intent from non-invasive biosignals is a central requirement for upper-limb myoelectric prostheses \cite{ref1}, \cite{ref2} and wearable human–machine interfaces (HMIs) \cite{ref3}. Surface electromyography (sEMG) captures superficial motor unit action potentials before visible limb motion \cite{ref2}, yet sEMG-only pattern recognition remains sensitive to changes in forearm orientation and gravitational loading \cite{ref4}, \cite{ref5}. Combining sEMG with tri-axial accelerometry (ACC) or inertial measurement units (IMUs) provides complementary macroscopic kinematics and gravitational posture alongside microscopic neuromuscular activation \cite{ref6}, \cite{ref7}, \cite{ref8}. This multimodal coupling is particularly valuable for transradial amputees, in whom residual muscle atrophy, scar tissue, and restricted socket geometry limit the number and spatial separability of recording sites \cite{ref6}, \cite{ref7}.

Despite high recognition accuracy in controlled laboratory sessions, wearable sEMG–inertial interfaces frequently suffer severe signal degradation during daily ambulatory wear \cite{ref5}, \cite{ref9}. In clinical prosthetic sockets and consumer-grade dry-electrode armbands, perspiration-induced impedance drift, socket pistoning during dynamic reaching, and cable fatigue all give rise to localized physical disturbances \cite{ref9}, \cite{ref10}, \cite{ref11}. Specifically, transient electrode lift-off breaks the skin–amplifier common-mode balance and injects high-amplitude $50/60\text{ Hz}$ powerline harmonics together with direct-current (DC) step offsets; mechanical skin traction produces non-stationary baseline wander (BW), motion artifacts (MOA), and saturation spikes; and intermittent wire or connector failures cause open-circuit channel dropout that zeros the affected signals \cite{ref10}, \cite{ref11}, \cite{ref12}. In practical wear, these failures rarely resemble an idealized, all-or-nothing loss of the entire sEMG modality \cite{ref13}. Instead, they appear as \emph{unknown partial channel corruption}, in which an arbitrary subset of $K$ out of $C_e$ sEMG electrodes degrades or drops out at unknown onset times while the remaining $C_e - K$ sEMG channels and a single forearm reference IMU serving as a posture reference stay intact \cite{ref9}, \cite{ref14}.

\begin{figure*}[!t]
\centering
\includegraphics[width=\textwidth]{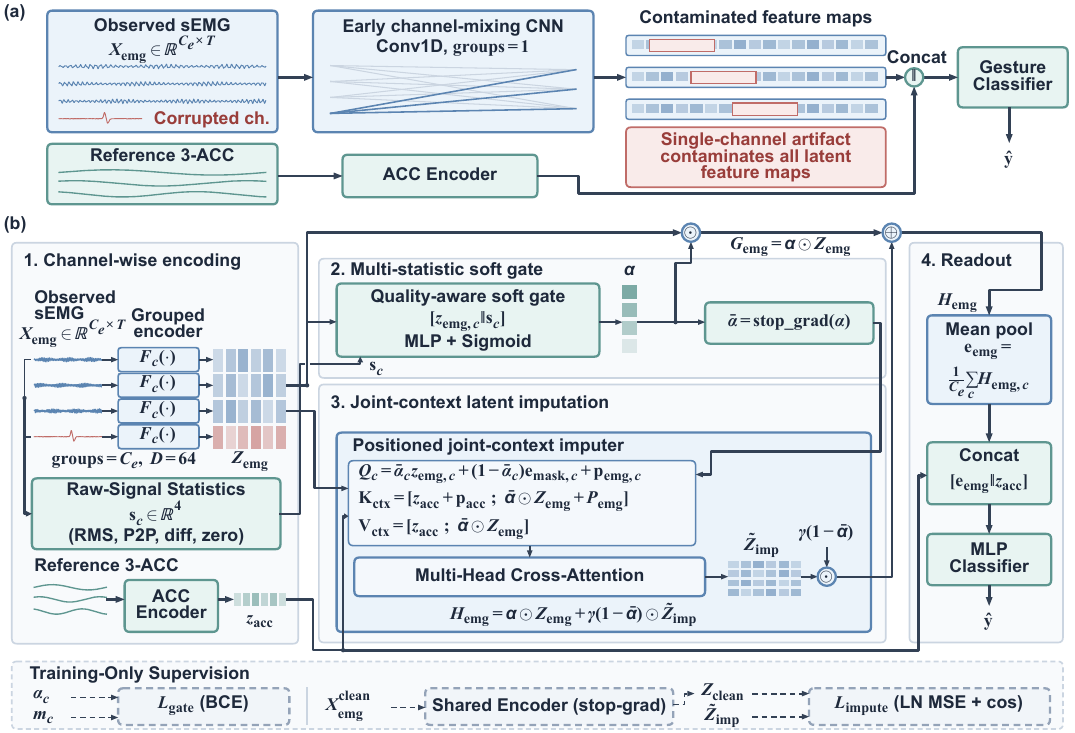}
\caption{Overview of early channel mixing and the proposed QA-CMIF framework. (a) Localized sEMG artifacts spread across latent feature maps through early channel mixing. (b) Channel-independent encoding, multi-statistic soft gating, and joint-context latent imputation with a confidence-weighted residual bypass.}
\label{fig:overview}
\end{figure*}

Unknown partial channel corruption exposes fundamental architectural limitations in existing multimodal hand gesture recognition (HGR) pipelines, as illustrated in Fig.~\ref{fig:overview}(a). Conventional multimodal fusion networks—from early channel-concatenation convolutional neural networks (CNNs) \cite{ref15}, \cite{ref16} and multi-feature attention backbones \cite{ref17}, \cite{ref18}, \cite{ref19} to hierarchical and incomplete-modality frameworks such as HyFusion \cite{ref20} and AiFusion \cite{ref13}—apply dense cross-channel convolutions in the first feature-extraction layer. Because every early filter sums over all $C_e$ input channels simultaneously, a high-amplitude lift-off shock or burst artifact on even two or three electrodes contaminates the entire latent feature map before any downstream fusion can intervene.

Two alternative strategies avoid this problem but have drawbacks of their own. The first pairs early channel-independent encoding with signal quality assessment (SQA) or soft reliability gating \cite{ref9}, \cite{ref12}. This prevents early cross-channel anomaly mixing, but the resulting gate-only attenuation merely suppresses corrupted channels toward zero. Zeroing out degraded channels discards the spatial myoelectric co-activation patterns recorded at the corresponding anatomical sites, which creates an information bottleneck when discriminating fine-grained 40- to 52-class finger and grasp taxonomies \cite{ref6}. The second strategy attempts to recover corrupted signals through recurrent incomplete-channel estimation \cite{ref10}, front-end time-domain waveform denoising \cite{ref11}, or masked waveform-reconstruction pre-training \cite{ref21}. However, reconstructing stochastic $2000\text{ Hz}$ sEMG waveforms incurs heavy encoder--decoder overhead and risks introducing synthetic waveform distortions under low signal-to-noise ratio (SNR) corruption, whereas gesture classification requires compact, discriminative latent representations rather than sample-level waveform generation \cite{ref22}.

To bridge the gap between early anomaly isolation and cross-channel context recovery, we propose Quality-Aware Cross-Modal Latent Imputation and Fusion (QA-CMIF), a lightweight end-to-end framework for sEMG--inertial HGR under unknown partial channel corruption shown in Fig.~\ref{fig:overview}(b). Rather than mixing channels at the input or reconstructing raw time-domain waveforms, QA-CMIF encodes each sEMG channel independently into a compact latent vector, termed a token, via grouped temporal convolutions and estimates a continuous reliability score for each channel with a soft gate driven by multiple signal statistics. It then repairs degraded channels in latent space through a cross-attention imputer whose queries carry learnable electrode-identity embeddings to retrieve information from both the reference inertial stream and the remaining high-confidence sEMG tokens, while a confidence-weighted residual bypass protects the intact channels. The main contributions of this work are as follows:

\begin{itemize}
\item \strong{Channel-independent encoding with continuous quality gating.} We design an encoding stage in which every sEMG channel is mapped to its own latent token without any cross-channel coupling, and combine it with a soft gate that estimates a continuous reliability score for each channel from four lightweight time-domain statistics and the latent token. The gate isolates localized high-amplitude artifacts before any spatial interaction and detects corrupted channels accurately, achieving an area under the receiver operating characteristic curve (AUROC) \ensuremath{\ge} 0.995 on unseen test repetitions.
\item \strong{Cross-modal latent imputation with explicit bad-channel supervision.} We introduce a cross-attention imputer that reconstructs the latent tokens of degraded channels from a joint context of the reference inertial token and the remaining reliable sEMG tokens. A confidence-weighted residual bypass preserves high-confidence channels through a mapping that approaches identity as $\alpha_c \rightarrow 1$, while an alignment loss computed strictly on corrupted channels after parameter-free layer normalization reduces the latent mean squared error of corrupted channels by 40\%–51\% relative to their gate-only attenuated representations within the same model.
\item \strong{Multi-cohort validation with a lightweight footprint.} We validate QA-CMIF on 29 subjects spanning transradial amputees from Ninapro DB3 recorded at $2000\text{ Hz}$, intact subjects from DB2 recorded with a medical-grade array at $2000\text{ Hz}$, and intact subjects from DB5 recorded with consumer armbands at $200\text{ Hz}$. Across 17-, 40-, and 52-class gesture sets, QA-CMIF consistently outperforms gate-only attenuation and corruption-augmented conventional fusion with only $0.175\text{M}$ parameters and $2.63\text{M}$ multiply–accumulate operations (MACs) at $2000\text{ Hz}$, while controlled ablations and degradation sweeps delineate its operating boundaries across signal regimes.
\end{itemize}

The remainder of this paper is organized as follows. Section II reviews related work, Section III presents the QA-CMIF methodology, Section IV describes the experimental setup, Section V reports the results, Section VI discusses the findings and limitations, and Section VII concludes the paper.

\section{Related Work}

\subsection{Multimodal sEMG–Inertial Hand Gesture Recognition}

Combining surface myoelectric signals with inertial measurements is a well-established strategy for mitigating limb-orientation sensitivity and improving multi-degree-of-freedom HGR \cite{ref4}, \cite{ref6}, \cite{ref7}. Early classifiers relied on concatenated handcrafted time–frequency sEMG and accelerometer descriptors \cite{ref6}, \cite{ref7}, and multi-view and multi-stream CNNs later learned joint spatiotemporal representations directly from windowed sensor streams \cite{ref15}, \cite{ref16}. More recently, hierarchical and attention-driven architectures have further improved decoding under clean conditions. Specifically, HyFusion \cite{ref20} integrates modality-specific and joint sEMG--ACC branches across multiple scales; MFN-SPSCA \cite{ref17}, MFF-Net \cite{ref18}, and NKDFF-CNN \cite{ref19} incorporate spatial partitioning and dual-view feature fusion; and recent \emph{IEEE JBHI} studies have extended multimodal fusion to biomechanical IMU–EMG attention \cite{ref23}, hierarchical EMG–FMG fusion (HEFF) \cite{ref8}, and contrastive meta-learning (STC+Meta) \cite{ref24}. Despite their strong nominal performance on clean recordings, these early-fusion and multi-branch backbones mix all $C_e$ sEMG channels at the first convolutional layer or across fixed spatial partitions, leaving the shared representation vulnerable to early cross-channel contagion when localized electrodes degrade in Fig.~\ref{fig:overview}(a).

\subsection{Signal Quality Assessment and Contaminant Mitigation in Myoelectric Interfaces}

Ambulatory sEMG recordings are prone to powerline interference (PLI), BW, MOA, and contact loss \cite{ref5}, \cite{ref27}, which has motivated extensive work on SQA and contaminant removal. Classical schemes flag corrupted segments using spectral energy ratios, variational mode decomposition, or one-class support vector machines \cite{ref25}, \cite{ref26}, \cite{ref30}, and Spanias \emph{et al.} \cite{ref9} showed that detecting and excluding disturbed EMG channels prevents catastrophic prosthetic control errors. Deep learning has since advanced contaminant scoring through the work of Machado et al. \cite{ref28}, Sauer et al. \cite{ref29}, and Prism-SQA \cite{ref12}. Other advances include incomplete-channel recurrent estimation with Att-RSV-LSTM \cite{ref10}, front-end waveform denoising with TrustEMG-Net \cite{ref11}, and masked-reconstruction foundation pre-training with TinyMyo \cite{ref21}. Unlike standalone quality scoring or computationally costly front-end waveform synthesis ahead of a classifier, QA-CMIF couples continuous multi-statistic soft gating directly with latent-space representation repair, recovering inter-channel muscle synergy cues \cite{ref31}, \cite{ref32} that hard channel rejection \cite{ref9} or gate-only attenuation discards.

\subsection{Incomplete Multimodal Fusion vs. Partial Channel Latent Imputation}

To address sensor unreliability in multimodal HGR, AiFusion \cite{ref13} introduced alignment-enhanced interactive fusion for complete and incomplete modality conditions, but it models incompleteness only at the whole-modality level, namely EMG-loss or ACC-loss, and retains dense early cross-channel convolutions within its sEMG encoder rather than addressing unknown partial channel corruption \cite{ref9}, \cite{ref14}. Meanwhile, latent-space feature completion and alignment have gained traction across biomedical informatics, including cross-modal sEMG--ACC latent compression \cite{ref22}, trait--state EEG imputation through TASTE \cite{ref33}, and semantic-guided latent completion through LFC-SATP-SGG \cite{ref34}, avoiding the overhead of raw signal generation. As summarized in Table I, QA-CMIF extends this paradigm to channel-level, fault-tolerant myoelectric decoding across both transradial amputee and intact cohorts.

\begin{table*}[!t]
\centering
\begingroup
\fontsize{8}{9.5}\selectfont
\setlength{\tabcolsep}{2pt}
\renewcommand{\arraystretch}{1.0}
\setlength{\TableUsable}{\dimexpr\textwidth-24pt\relax}
\caption{OBJECTIVE COMPARISON OF REPRESENTATIVE MYOELECTRIC FUSION, DENOISING, AND IMPUTATION PARADIGMS}
\label{tab:I}
\begin{tabular}{@{}>{\raggedright\arraybackslash}p{0.140000\TableUsable}>{\raggedright\arraybackslash}p{0.085000\TableUsable}>{\raggedright\arraybackslash}p{0.145000\TableUsable}>{\raggedright\arraybackslash}p{0.200000\TableUsable}>{\raggedright\arraybackslash}p{0.140000\TableUsable}>{\raggedright\arraybackslash}p{0.140000\TableUsable}>{\raggedright\arraybackslash}p{0.150000\TableUsable}@{}}
\toprule
\TCell{I}{0}{0}{Method / Paradigm} & \TCell{I}{0}{1}{Input Modalities} & \TCell{I}{0}{2}{Target Degradation Granularity} & \TCell{I}{0}{3}{Processing \& Repair Domain} & \TCell{I}{0}{4}{Explicit Bad-Channel Latent Supervision} & \TCell{I}{0}{5}{Transradial Amputee Validation} & \TCell{I}{0}{6}{Matched $2000\text{ Hz}$ Params / MACs $^{\text{a}}$} \\
\midrule
\TCell{I}{1}{0}{HyFusion \cite{ref20} (\emph{IEEE Sensors J.} 2023)} & \TCell{I}{1}{1}{sEMG + ACC} & \TCell{I}{1}{2}{Nominal clean recordings} & \TCell{I}{1}{3}{Multi-branch features with dense first-layer cross-channel mixing} & \TCell{I}{1}{4}{No} & \TCell{I}{1}{5}{Yes (DB3, DB7; clean only)} & \TCell{I}{1}{6}{$390.5\text{K} \;/\; 13.75\text{M}$ (HyFusion-Lite)} \\
\midrule
\TCell{I}{2}{0}{MFF-Net \cite{ref18} (\emph{IEEE JBHI} 2025)} & \TCell{I}{2}{1}{sEMG only} & \TCell{I}{2}{2}{Sparse-channel clean / transfer} & \TCell{I}{2}{3}{Time–frequency–spatial mixed feature fusion} & \TCell{I}{2}{4}{No} & \TCell{I}{2}{5}{Yes (DB3 subset)} & \TCell{I}{2}{6}{— (Unimodal sEMG)} \\
\midrule
\TCell{I}{3}{0}{HEFF \cite{ref8} (\emph{IEEE JBHI} 2026)} & \TCell{I}{3}{1}{sEMG + FMG} & \TCell{I}{3}{2}{Cross-day \& limb-position drift} & \TCell{I}{3}{3}{Data/feature/decision pyramid fusion} & \TCell{I}{3}{4}{No} & \TCell{I}{3}{5}{No (15 intact subjects)} & \TCell{I}{3}{6}{— (Custom 8-ch bracelet)} \\
\midrule
\TCell{I}{4}{0}{TrustEMG-Net \cite{ref11} (\emph{IEEE JBHI} 2025)} & \TCell{I}{4}{1}{Single-ch sEMG} & \TCell{I}{4}{2}{Waveform noise (PLI, BW, ECG, MOA)} & \TCell{I}{4}{3}{Time-domain waveform reconstruction (U-Net + Transformer)} & \TCell{I}{4}{4}{No (Waveform $L_1$ loss)} & \TCell{I}{4}{5}{No (Intact DB2)} & \TCell{I}{4}{6}{— (Front-end denoiser)} \\
\midrule
\TCell{I}{5}{0}{Prism-SQA \cite{ref12} (\emph{IEEE JBHI} 2026)} & \TCell{I}{5}{1}{Single-ch sEMG} & \TCell{I}{5}{2}{Multi-contaminant waveform distortion} & \TCell{I}{5}{3}{Waveform source separation + fingerprint verification} & \TCell{I}{5}{4}{No (Quality score target)} & \TCell{I}{5}{5}{Yes (DB3 SQA test)} & \TCell{I}{5}{6}{— (Standalone SQA network)} \\
\midrule
\TCell{I}{6}{0}{AiFusion \cite{ref13} (\emph{IEEE TNSRE} 2023)} & \TCell{I}{6}{1}{sEMG + ACC/IMU} & \TCell{I}{6}{2}{Complete modality absence (EMG-loss / ACC-loss)} & \TCell{I}{6}{3}{Dense first-layer cross-channel mixing + interactive alignment} & \TCell{I}{6}{4}{No} & \TCell{I}{6}{5}{Yes (DB3 5-subject subset)} & \TCell{I}{6}{6}{$376.1\text{K} \;/\; 9.90\text{M}$ (AiFusion-Lite)} \\
\midrule
\TCell{I}{7}{0}{Vanilla-Aug (Control)} & \TCell{I}{7}{1}{sEMG + ACC} & \TCell{I}{7}{2}{Unknown partial channel corruption} & \TCell{I}{7}{3}{Dense first-layer cross-channel mixing} & \TCell{I}{7}{4}{No} & \TCell{I}{7}{5}{Yes (DB3, $n=9$ (17-class) and $n=6$ (40-class))} & \TCell{I}{7}{6}{$174.5\text{K} \;/\; 4.07\text{M}$} \\
\midrule
\TCell{I}{8}{0}{Gate-Only (Control)} & \TCell{I}{8}{1}{sEMG + ACC} & \TCell{I}{8}{2}{Unknown partial channel corruption} & \TCell{I}{8}{3}{Channel-independent early encoding + soft-gate attenuation} & \TCell{I}{8}{4}{No (Gate BCE $\mathcal{L}_{\text{gate}}$ only)} & \TCell{I}{8}{5}{Yes (DB3, $n=9$ (17-class) and $n=6$ (40-class))} & \TCell{I}{8}{6}{$139.8\text{K} \;/\; 2.21\text{M}$} \\
\midrule
\TCell{I}{9}{0}{\strong{QA-CMIF (Ours)}} & \TCell{I}{9}{1}{\strong{sEMG + ACC}} & \TCell{I}{9}{2}{\strong{Unknown partial channel corruption \& modality dropout}} & \TCell{I}{9}{3}{\strong{Channel-independent early encoding + soft gate + residual bypass + joint-context latent imputation ($D=64$)}} & \TCell{I}{9}{4}{\strong{Yes (Parameter-free LayerNorm $\mathcal{L}_{\text{impute}}$ on bad tokens)}} & \TCell{I}{9}{5}{\strong{Yes (DB3, $n=9$ (17-class) and $n=6$ (40-class))}} & \TCell{I}{9}{6}{\strong{$174.9\text{K} \;/\; 2.63\text{M}$}} \\
\bottomrule
\end{tabular}
\par\vspace{3pt}
\begin{minipage}{\textwidth}\footnotesize
\TNote{I}{0}{$^{\text{a}}$ Parameter counts and per-window multiply–accumulate operations (MACs) are reported for models evaluated under the unified $2000\text{ Hz}$ protocol ($12\text{-EMG} + 3\text{-ACC}$, $T=400$, 17 classes); HyFusion-Lite and AiFusion-Lite denote lightweight adaptations that use the same single tri-axial accelerometer input and are trained under the identical six-condition corruption augmentation.}\par
\end{minipage}
\endgroup
\end{table*}

\section{Methodology}

As illustrated in Fig.~\ref{fig:overview}(a), conventional multimodal sEMG--inertial networks apply dense cross-channel convolutions with one filter group across all $C_e$ electrodes at the initial encoding stage, allowing high-variance artifacts on even a few degraded channels to propagate throughout the shared latent representation. To overcome this vulnerability, the proposed QA-CMIF framework in Fig.~\ref{fig:overview}(b) restructures the pipeline into four sequential stages coupled by a multi-task training objective, comprising 1) channel-independent temporal encoding through grouped convolutions with $C_e$ filter groups alongside a separate inertial encoder, 2) multi-statistic continuous quality-aware soft gating, 3) confidence-weighted residual bypass and joint-context latent imputation, and 4) bad-channel alignment supervision with parameter-free layer normalization.

\subsection{Problem Formulation and Inference-Time Observability}

Consider a synchronized multimodal sliding window of physical duration $200\text{ ms}$ with a $50\text{ ms}$ stride \cite{ref1}, \cite{ref3}. Each window comprises a multi-channel sEMG recording $X_{\text{emg}} = [X_{\text{emg},1}, \dots, X_{\text{emg},C_e}]^\top \in \mathbb{R}^{C_e \times T}$ from $C_e$ electrodes and a pre-synchronized three-axis acceleration stream $X_{\text{acc}} \in \mathbb{R}^{C_a \times T}$ with $C_a = 3$, denoted 3-ACC, from a single forearm reference IMU \cite{ref3}, \cite{ref6}, \cite{ref14}. Under $2000\text{ Hz}$ raw-sEMG acquisitions \cite{ref6}, each window spans $T = 400$ samples across $C_e = 12$ electrodes; under $200\text{ Hz}$ pre-conditioned dry-electrode armbands \cite{ref3}, each window spans $T = 40$ samples across $C_e = 16$ channels. All channels are $z$-score standardized per subject using training-split statistics, and the network outputs gesture logits $\hat{\mathbf{y}} \in \mathbb{R}^{M}$ to predict the discrete hand gesture category $\hat{y}_{\mathrm{cls}} = \arg\max_{j \in \{1,\dots,M\}} \hat{y}_j$ across $M \in \{17, 40, 52\}$ active gesture classes.

To formalize ambulatory electrode degradation without violating deployment constraints, we enforce a strict boundary between training supervision and inference-time observability in Fig.~\ref{fig:overview}(b). During training, each clean window $(X_{\text{emg}}^{\text{clean}}, X_{\text{acc}}^{\text{clean}})$ is paired with an on-the-fly corrupted window $(X_{\text{emg}}, X_{\text{acc}}) = \mathcal{T}_{\text{corrupt}}(X_{\text{emg}}^{\text{clean}}, X_{\text{acc}}^{\text{clean}}; \mathbf{m})$, where $\mathbf{m} = [m_1, \dots, m_{C_e}]^\top \in \{0, 1\}^{C_e}$ is the ground-truth binary sEMG channel-health indicator. Here, $m_c = 1$ if intact and $m_c = 0$ if corrupted, while $X_{\text{acc}} = X_{\text{acc}}^{\text{clean}}$ in all conditions except whole-modality inertial dropout. Importantly, $\mathbf{m}$ and $X_{\text{emg}}^{\text{clean}}$ are used exclusively during training to compute the auxiliary losses $\mathcal{L}_{\text{gate}}$ and $\mathcal{L}_{\text{impute}}$. At inference time, the network receives only the observed window $(X_{\text{emg}}, X_{\text{acc}})$; the health mask $\mathbf{m}$, the corrupted channel count $K = \sum_{c=1}^{C_e}(1 - m_c)$, and the corruption type are completely unknown.

\subsection{Channel-Independent Temporal Encoding with Sampling-Rate-Specific Front Ends}

To prevent cross-channel artifact leakage during initial temporal feature extraction, QA-CMIF encodes $X_{\text{emg}}$ through a grouped 1-D convolutional backbone with $C_e$ filter groups, terminated by token-wise layer normalization along the feature dimension $D = 64$ without cross-channel batch normalization in Fig.~\ref{fig:overview}(b). Each electrode waveform $X_{\text{emg},c} \in \mathbb{R}^{1 \times T}$ is mapped by a channel-specific temporal operator $\mathcal{F}_c(\cdot)$ into a latent token $z_{\text{emg},c} = \mathcal{F}_c(X_{\text{emg},c}) \in \mathbb{R}^D$, forming $Z_{\text{emg}} = [z_{\text{emg},1}, \dots, z_{\text{emg},C_e}]^\top \in \mathbb{R}^{C_e \times D}$. Because neither grouped convolutions nor per-token LayerNorm couples distinct electrodes $c \neq c'$, every cross-channel partial derivative vanishes within the temporal encoder:

\begin{equation}
\frac{\partial z_{\text{emg},c'}}{\partial X_{\text{emg},c}} = \mathbf{0}, \quad \forall c' \neq c
\label{eq:1}
\end{equation}

Equation (1) ensures that within the temporal encoder itself, arbitrary corruption on input channel $c$ leaves the initial latent tokens $\{z_{\text{emg},c'}\}_{c' \neq c}$ of all remaining intact channels invariant prior to quality gating and cross-channel fusion. In parallel, the three-axis acceleration stream $X_{\text{acc}} \in \mathbb{R}^{3 \times T}$ is encoded by a three-stage single-group 1-D CNN with feature widths of $64 \rightarrow 128 \rightarrow 64$ into a single global kinematic posture token $z_{\text{acc}} \in \mathbb{R}^D$.

To accommodate the $10\times$ sampling-rate difference between $2000\text{ Hz}$ raw-sEMG arrays and $200\text{ Hz}$ pre-conditioned armbands while keeping the downstream token geometry identical, only the first temporal down-sampling stage is adapted to the acquisition regime, while the three grouped convolution stages expand each electrode through per-channel feature widths of $8 \rightarrow 16 \rightarrow 64$. In $2000\text{ Hz}$ recordings with $T = 400$, surface motor unit action potential waveforms typically span $5\text{--}15\text{ ms}$ \cite{ref2}. The $2000\text{ Hz}$ encoder therefore begins with a wide-kernel strided convolution using kernel size $k = 25$, a $12.5\text{ ms}$ receptive field, stride $s = 5$, and one filter group per electrode, followed by GELU activation and max-pooling by a factor of 2, compressing the sequence from 400 to 80 and then to 40 samples. Two further strided grouped convolution blocks with $k = 5, s = 2$ and $k = 3, s = 2$, global average pooling, and layer normalization over $D = 64$ features complete the encoder along the sequence-length trajectory $400 \rightarrow 80 \rightarrow 40 \rightarrow 20 \rightarrow 10 \rightarrow 1$.

For the $200\text{ Hz}$ Myo armbands with $T = 40$ \cite{ref3}, the same $10\times$ down-sampling front end would collapse the sequence to 4 samples at the first layer. The $200\text{ Hz}$ encoder therefore omits it and applies three grouped convolution blocks with strides $1, 2,$ and $2$, producing the sequence-length trajectory $40 \rightarrow 40 \rightarrow 20 \rightarrow 10 \rightarrow 1$, so that both regimes share the same terminal resolution and the same latent token shape $Z_{\text{emg}} \in \mathbb{R}^{C_e \times 64}$.

\subsection{Multi-Statistic Continuous Quality-Aware Soft Gating}

Because LayerNorm normalizes feature vectors to unit variance across $D$, the latent token $z_{\text{emg},c}$ alone discards absolute amplitude scale and raw sample continuity, making it difficult to distinguish a normalized high-voltage lift-off transient from a vigorous contraction or a dropout flatline from true rest. To preserve physical scale and continuity cues without the computational cost of time–frequency transforms \cite{ref12}, \cite{ref28}, we extract four lightweight time-domain descriptors $\mathbf{s}_c = [s_{c,1}, s_{c,2}, s_{c,3}, s_{c,4}]^\top \in \mathbb{R}^4$ from each standardized waveform $X_{\text{emg},c} = (X_{c,1}, \dots, X_{c,T})$:

\begin{equation}
\begin{gathered}s_{c,1} = \log\left(\max\left(\sqrt{\frac{1}{T}\sum_{t=1}^{T} X_{c,t}^2},\; \epsilon\right)\right),\\[-1pt] \quad \epsilon = 10^{-6}\end{gathered}
\label{eq:2}
\end{equation}

\begin{equation}
s_{c,2} = \log\left(1 + \left(\max_{1 \le t \le T} X_{c,t} - \min_{1 \le t \le T} X_{c,t}\right)\right)
\label{eq:3}
\end{equation}

\begin{equation}
s_{c,3} = \log\left(1 + \frac{1}{T-1}\sum_{t=1}^{T-1} \big|X_{c,t+1} - X_{c,t}\big|\right)
\label{eq:4}
\end{equation}

\begin{equation}
s_{c,4} = \frac{1}{T}\sum_{t=1}^{T} \mathbb{I}(X_{c,t} = 0)
\label{eq:5}
\end{equation}

where $\mathbb{I}(\cdot)$ is the indicator function. These four descriptors are designed to provide complementary sensitivity to amplitude, derivative, and dropout anomalies. Specifically, the clamped log root-mean-square energy $s_{c,1}$ and log peak-to-peak range $s_{c,2}$ track broadband amplitude shifts and impulsive shocks across orders of magnitude; the log mean absolute first-order difference $s_{c,3}$ responds to rapid high-frequency oscillations and step transients; and the exact zero-sample proportion $s_{c,4}$ flags digital channel zeroing, or dropout.

For each channel $c$, $[z_{\text{emg},c} \,\|\, \mathbf{s}_c] \in \mathbb{R}^{D+4}$ is fed into a shared two-layer perceptron $\phi_{\text{gate}}: \mathbb{R}^{68} \rightarrow \mathbb{R}$ with $68 \rightarrow 32 \rightarrow 1$ units and GELU activation, and a sigmoid function $\sigma(\cdot)$ converts its output into a continuous reliability score:

\begin{equation}
\begin{gathered}g_c = \phi_{\text{gate}}\big([z_{\text{emg},c} \,\|\, \mathbf{s}_c]\big),\\ \alpha_c = \sigma(g_c) \in (0, 1)\end{gathered}
\label{eq:6}
\end{equation}

Stacking $\boldsymbol{\alpha} = [\alpha_1, \dots, \alpha_{C_e}]^\top \in (0, 1)^{C_e \times 1}$ yields the soft-gated representation $G_{\text{emg}} = \boldsymbol{\alpha} \odot Z_{\text{emg}} \in \mathbb{R}^{C_e \times D}$, where $\odot$ denotes row-wise broadcast multiplication in Fig.~\ref{fig:overview}(b). Note that $\alpha_c$ operates throughout forward propagation as a differentiable soft weight rather than a hard binary switch; a threshold of $\alpha_c \ge 0.5$ is used solely for post-hoc diagnostic reporting.

\subsection{Confidence-Weighted Residual Bypass and Joint-Context Latent Imputation}

While the Gate-Only soft-gating model suppresses high-variance noise injection, attenuating $K$ corrupted channels toward zero dilutes forearm spatial co-activation patterns. To reconstruct degraded channel representations in latent space while decoupling the gate parameters from imputation-branch gradients, we detach the reliability weights via $\bar{\alpha}_c = \text{stop\_grad}(\alpha_c)$ inside the imputation branch in Fig.~\ref{fig:overview}(b). For each electrode $c$, we form a query vector $Q_c \in \mathbb{R}^D$ by interpolating between $z_{\text{emg},c}$ and a channel-specific learnable mask token $\mathbf{e}_{\text{mask},c} \in \mathbb{R}^D$, augmented with a learnable electrode-identity embedding $\mathbf{p}_{\text{emg},c} \in \mathbb{R}^D$:

\begin{equation}
\begin{gathered}Q_c = \bar{\alpha}_c z_{\text{emg},c} + (1 - \bar{\alpha}_c)\mathbf{e}_{\text{mask},c} + \mathbf{p}_{\text{emg},c},\\ Q = [Q_1, \dots, Q_{C_e}]^\top \in \mathbb{R}^{C_e \times D}\end{gathered}
\label{eq:7}
\end{equation}

When channel $c$ is severely corrupted such that $\bar{\alpha}_c \rightarrow 0$, $Q_c$ reduces to $\mathbf{e}_{\text{mask},c} + \mathbf{p}_{\text{emg},c}$, prompting the imputer to query the context pool for the expected feature at anatomical site $c$.

With the single reference accelerometer in the 3-ACC configuration, the imputer retrieves complementary cues from a joint context pool of $1 + C_e$ tokens that couples macroscopic limb posture with intact inter-muscular spatial synergy. The pool comprises the global inertial token $z_{\text{acc}} \in \mathbb{R}^D$ and the reliability-attenuated sEMG tokens $\bar{\boldsymbol{\alpha}} \odot Z_{\text{emg}} \in \mathbb{R}^{C_e \times D}$ \cite{ref7}, \cite{ref31}. In our default configuration, modality and channel identity embeddings, $\mathbf{p}_{\text{acc}} \in \mathbb{R}^{1 \times D}$ and $P_{\text{emg}} \in \mathbb{R}^{C_e \times D}$, are added to the keys $\mathbf{K}_{\text{ctx}}$ so that each context token can be addressed by its modality or electrode identity, while the values $\mathbf{V}_{\text{ctx}}$ remain free of positional bias:

\begin{equation}
\begin{gathered}\mathbf{K}_{\text{ctx}} = \big[\, z_{\text{acc}} + \mathbf{p}_{\text{acc}} \;;\; \bar{\boldsymbol{\alpha}} \odot Z_{\text{emg}} + P_{\text{emg}} \,\big]\\[-2pt] \in \mathbb{R}^{(1 + C_e) \times D},\\ \mathbf{V}_{\text{ctx}} = \big[\, z_{\text{acc}} \;;\; \bar{\boldsymbol{\alpha}} \odot Z_{\text{emg}} \,\big]\\[-2pt] \in \mathbb{R}^{(1 + C_e) \times D}\end{gathered}
\label{eq:8}
\end{equation}

Multi-head cross-attention (MHA) with $H = 4$ heads and $d_k = 16$ followed by a residual layer normalization (LN) and a two-layer feedforward network (FFN) with $64 \rightarrow 128 \rightarrow 64$ units, GELU activation, and dropout of 0.2 produces the imputed tokens $\tilde{Z}_{\text{imp}} \in \mathbb{R}^{C_e \times D}$:

\begin{equation}
\begin{gathered}A_{\text{attn}} = \text{LN}\big(Q + \text{MHA}(Q, \mathbf{K}_{\text{ctx}}, \mathbf{V}_{\text{ctx}})\big),\\ \tilde{Z}_{\text{imp}} = \text{LN}\big(A_{\text{attn}} + \text{FFN}(A_{\text{attn}})\big)\end{gathered}
\label{eq:9}
\end{equation}

Routing all channels unconditionally through cross-attention by setting $H_{\text{emg}} = \tilde{Z}_{\text{imp}}$ would replace intact electrode tokens with context-averaged outputs and smooth out channel-specific neuromuscular details. To preserve high-confidence channel representations while selectively repairing degraded channels, QA-CMIF combines the gated direct pathway and the imputation pathway through a confidence-weighted residual bypass scaled by a learnable scalar $\gamma \in \mathbb{R}$ in Fig.~\ref{fig:overview}(b):

\begin{equation}
H_{\text{emg}} = \boldsymbol{\alpha} \odot Z_{\text{emg}} + \gamma \cdot (\mathbf{1} - \bar{\boldsymbol{\alpha}}) \odot \tilde{Z}_{\text{imp}} \in \mathbb{R}^{C_e \times D}
\label{eq:10}
\end{equation}

Under (10), active gate gradients flow only through the direct pathway $\boldsymbol{\alpha} \odot Z_{\text{emg}}$ while $\mathbf{1} - \bar{\boldsymbol{\alpha}}$ remains detached; consequently, an intact channel with $\alpha_c \rightarrow 1$ retains its distinctive token $z_{\text{emg},c}$, whereas a corrupted channel with $\alpha_c \rightarrow 0$ is replaced by $\gamma \tilde{z}_{\text{imp},c}$. The repaired tokens $H_{\text{emg}}$ are mean-pooled across channels and concatenated with $z_{\text{acc}}$ to predict the gesture logits $\hat{\mathbf{y}} \in \mathbb{R}^M$ via a two-layer MLP classifier with $128 \rightarrow 128 \rightarrow M$ units, GELU activation, and dropout of 0.2:

\begin{equation}
\begin{gathered}\mathbf{e}_{\text{emg}} = \frac{1}{C_e}\sum_{c=1}^{C_e} H_{\text{emg},c} \in \mathbb{R}^D,\\ \hat{\mathbf{y}} = \text{MLP}_{\text{cls}}\big([\mathbf{e}_{\text{emg}} \,\|\, z_{\text{acc}}]\big) \in \mathbb{R}^M\end{gathered}
\label{eq:11}
\end{equation}

In our $2000\text{ Hz}$ array pipeline, QA-CMIF employs the imputer configuration in (8) with identity embeddings on the keys and an initial bypass scale of $\gamma_0 = 0.5$. In the $200\text{ Hz}$ 16-channel armband pipeline, where pre-conditioned envelope streams carry lower per-channel temporal resolution across a denser circumferential ring, identity embeddings are added only to the queries so that $\mathbf{K}_{\text{ctx}} = \mathbf{V}_{\text{ctx}} = [z_{\text{acc}} \;;\; \bar{\boldsymbol{\alpha}} \odot Z_{\text{emg}}]$, paired with a more conservative initial scale of $\gamma_0 = 0.1$ to stabilize early cross-attention updates, with $\gamma$ optimized end-to-end without manual box constraints in both regimes.

\subsection{Multi-Task Objective with Exact Bad-Channel Alignment Loss}

Without explicit latent reconstruction supervision, the imputer receives only task-level classification gradients diluted by channel pooling in (11). During training, we obtain clean target tokens by passing the paired uncorrupted window $X_{\text{emg}}^{\text{clean}}$ through the shared encoder with stopped gradients, $Z_{\text{clean}} = \text{stop\_grad}\big(E_{\text{emg}}(X_{\text{emg}}^{\text{clean}})\big) \in \mathbb{R}^{C_e \times D}$. However, directly penalizing unnormalized Euclidean distance across all channels encourages the encoder to shrink token norms toward zero and wastes capacity copying already intact channels for which $m_{bc} = 1$.

To enforce scale-invariant alignment strictly on corrupted channels while preventing token norm collapse, we apply parameter-free, affine-free layer normalization $\text{LN}(\cdot)$ along $D = 64$ to obtain $\mathbf{u}_{bc} = \text{LN}(\tilde{z}_{\text{imp},bc})$ and $\mathbf{v}_{bc} = \text{LN}(z_{\text{clean},bc})$. Because affine-free layer normalization projects each token onto a zero-mean hypersphere of radius $\sqrt{D}$ where $\frac{1}{D}\|\mathbf{u}_{bc}-\mathbf{v}_{bc}\|_2^2 = 2\big(1-\cos(\mathbf{u}_{bc},\mathbf{v}_{bc})\big)$, summing the normalized squared Euclidean distance and the cosine distance over the corrupted token set $\mathcal{B} = \{(b, c) \mid m_{bc} = 0\}$ of size $N_{\mathcal{B}} = |\mathcal{B}|$ forms a unified scale-invariant directional alignment penalty on the hypersphere when $N_{\mathcal{B}} > 0$, whereas $\mathcal{L}_{\text{impute}} = 0$ when $N_{\mathcal{B}} = 0$, as on clean or inertial-dropout windows:

\begin{equation}
\begin{aligned}
&\mathcal{L}_{\text{impute}} =\\[-2pt]
&\begin{cases}
\begin{aligned}
&\displaystyle\frac{1}{N_{\mathcal{B}}}\sum_{(b,c)\in\mathcal{B}}
 \left[\frac{1}{D}\big\|\mathbf{u}_{bc}-\mathbf{v}_{bc}\big\|_2^2\right.\\
&\qquad\left.{}+\Big(1-\cos(\mathbf{u}_{bc},\mathbf{v}_{bc})\Big)\right],
\end{aligned}&N_{\mathcal{B}}>0,\\[4pt]
0,&N_{\mathcal{B}}=0
\end{cases}
\end{aligned}
\label{eq:12}
\end{equation}

Each training minibatch is evaluated on both the clean window $(X_{\text{emg}}^{\text{clean}}, X_{\text{acc}}^{\text{clean}})$ and its corrupted counterpart $(X_{\text{emg}}, X_{\text{acc}})$. Combining the clean and corrupted cross-entropy losses $\mathcal{L}_{\text{CE}}^{\text{clean}}$ and $\mathcal{L}_{\text{CE}}^{\text{corrupt}}$ with the binary cross-entropy gate loss $\mathcal{L}_{\text{gate}} = \frac{1}{B C_e}\sum_{b,c} \text{BCE}\big(\sigma(g_{bc}), m_{bc}\big)$, the composite training objective is:

\begin{equation}
\mathcal{L}_{\text{total}} = \mathcal{L}_{\text{CE}}^{\text{clean}} + \mathcal{L}_{\text{CE}}^{\text{corrupt}} + \lambda_{\text{gate}}\mathcal{L}_{\text{gate}} + \lambda_{\text{impute}}\mathcal{L}_{\text{impute}}
\label{eq:13}
\end{equation}

where $\lambda_{\text{gate}} = 0.5$ and $\lambda_{\text{impute}} = 0.2$ are fixed across all cohorts. Because $\mathbf{m}$ and $X_{\text{emg}}^{\text{clean}}$ are generated automatically on the fly during training, QA-CMIF requires no manual clinical quality annotations.

\section{Experimental Setup}

\subsection{Datasets, Subject Cohorts, and Label Selection}

We evaluated QA-CMIF across three public Ninapro benchmarks spanning clinical transradial amputees, high-rate medical arrays, and low-rate consumer armbands under a unified single-reference inertial setup, termed 3-ACC, that uses the first tri-axial accelerometer stream, as summarized in Table II \cite{ref3}, \cite{ref6}, \cite{ref14}; the 36-channel alternative is examined in Section VI-B. Throughout the paper, all tasks classify active movement categories while excluding the inter-trial rest state carrying label 0, where B17 denotes Exercise B with 17 wrist and hand classes, BC40 denotes Exercises B and C with 40 grasping and wrist classes, and ABC52 denotes Exercises A, B, and C with 52 finger, wrist, and grasping classes. On Ninapro DB3, recorded using Delsys Trigno sensors at $2000\text{ Hz}$ with 12-EMG and 3-ACC inputs \cite{ref6}, 11 transradial amputees, s1--s11, were enrolled, as detailed in Supplementary Table S1. Subjects s6, with $40\%$ remaining forearm, and s7, with $0\%$ remaining forearm, contained only 10 active sEMG channels because EMG columns 9 and 10 were identically zero, and were excluded a priori under our 12-effective-channel criterion. The remaining nine 12-channel amputees form the primary clinical cohort DB3-All9 with $n = 9$ on B17, which is partitioned into the pre-specified six-amputee subset DB3-Core6 with $n = 6$, evaluated on both B17 and BC40, and the three-amputee confirmation subset DB3-Holdout3 with $n = 3$ on B17. Across all datasets, Holdout subsets denote participants reserved for zero-retuning confirmation after locking architectural hyperparameters on the corresponding development splits.

For intact-limb validation, we evaluated the first 10 intact subjects, s1--s10, from Ninapro DB2, recorded using Delsys Trigno at $2000\text{ Hz}$ with 12-EMG and 3-ACC inputs \cite{ref6}, forming DB2-All10 with $n = 10$ on B17, alongside all 10 intact subjects, s1--s10, from Ninapro DB5, recorded using dual Thalmic Myo armbands at $200\text{ Hz}$ with 16-EMG and 3-ACC inputs \cite{ref3}, forming DB5-All10 with $n = 10$ on both B17 and ABC52. As listed in Table II, both 10-subject cohorts are partitioned into five-subject development subsets, DB2-Dev5 and DB5-Dev5 for s1--s5, and five-subject confirmation subsets, DB2-Holdout5 and DB5-Holdout5 for s6--s10.

All experiments followed a strict within-subject, cross-repetition split using official repetitions $\{1, 3, 4, 6\}$ for training and repetitions $\{2, 5\}$ for testing \cite{ref3}, \cite{ref6}. Recordings were segmented within constant-label, constant-repetition intervals using a $200\text{ ms}$ window and $50\text{ ms}$ stride, ensuring zero raw-sample overlap between splits, with per-channel $z$-score statistics fitted on the training split only. Segmentation used the official movement-aligned fields restimulus and rerepetition after verifying that all target classes were present in both splits across all evaluated cohorts; amputee s1 lacked complete Exercise C recordings and was therefore evaluated only on B17, as documented in Supplementary Table S1.

\begin{table*}[!t]
\centering
\begingroup
\fontsize{8}{9.5}\selectfont
\setlength{\tabcolsep}{2pt}
\renewcommand{\arraystretch}{1.0}
\setlength{\TableUsable}{\dimexpr\textwidth-20pt\relax}
\caption{SUMMARY OF EXPERIMENTAL BENCHMARKS, COHORT DEFINITIONS, ACQUISITION SETUPS, AND PROTOCOL PARAMETERS}
\label{tab:II}
\begin{tabular}{@{}>{\raggedright\arraybackslash}p{0.170000\TableUsable}>{\raggedright\arraybackslash}p{0.140000\TableUsable}>{\raggedright\arraybackslash}p{0.170000\TableUsable}>{\raggedright\arraybackslash}p{0.200000\TableUsable}>{\raggedright\arraybackslash}p{0.140000\TableUsable}>{\raggedright\arraybackslash}p{0.180000\TableUsable}@{}}
\toprule
\TCell{II}{0}{0}{Benchmark \& Population} & \TCell{II}{0}{1}{Sensor Hardware \& Sampling Rate} & \TCell{II}{0}{2}{Input Channels \& Window ($200\text{ ms} / 50\text{ ms}$)} & \TCell{II}{0}{3}{Cohort / Subset Name ($n$, Subject IDs)} & \TCell{II}{0}{4}{Evaluated Tasks (Classes)} & \TCell{II}{0}{5}{QA-CMIF Parameterization $^{\text{a}}$} \\
\midrule
\TCell{II}{1}{0}{\strong{Ninapro DB3} \cite{ref6}\newline Clinical Transradial Amputees\newline \emph{(11 enrolled; s6, s7 excluded due to 10 effective EMG channels)}} & \TCell{II}{1}{1}{Delsys Trigno Wireless Array\newline  $f_s = 2000\text{ Hz}$ \newline (Raw surface EMG)} & \TCell{II}{1}{2}{$12\text{-EMG} + 3\text{-ACC}$ \newline  $T = 400$, stride $100$ \newline Train: rep $\{1,3,4,6\}$ \newline Test: rep $\{2,5\}$} & \TCell{II}{1}{3}{DB3-All9 ($n = 9$: s1–s5, s8–s11)\newline DB3-Core6 ($n = 6$: s2, s4, s5, s8, s9, s11) $^{\text{b}}$ \newline DB3-Holdout3 ($n = 3$: s1, s3, s10) $^{\text{b}}$} & \TCell{II}{1}{4}{B17 (17 classes)\newline B17 (17) \& BC40 (40)\newline B17 (17 classes)} & \TCell{II}{1}{5}{Front end for $2000\text{ Hz}$ ($k=25$, $s=5$, max-pool 2)\newline Identity embeddings on keys ($\gamma_0 = 0.5$)} \\
\midrule
\TCell{II}{2}{0}{\strong{Ninapro DB2} \cite{ref6}\newline Intact-Limb Medical Array} & \TCell{II}{2}{1}{Delsys Trigno Wireless Array\newline  $f_s = 2000\text{ Hz}$ \newline (Raw surface EMG)} & \TCell{II}{2}{2}{$12\text{-EMG} + 3\text{-ACC}$ \newline  $T = 400$, stride $100$ \newline Train: rep $\{1,3,4,6\}$ \newline Test: rep $\{2,5\}$} & \TCell{II}{2}{3}{DB2-All10 ($n = 10$: s1–s10)\newline DB2-Holdout5 ($n = 5$: s6–s10) $^{\text{b}}$ \newline DB2-Dev5 ($n = 5$: s1–s5) $^{\text{b}}$} & \TCell{II}{2}{4}{B17 (17 classes)\newline B17 (17 classes)\newline B17 (17 classes)} & \TCell{II}{2}{5}{Front end for $2000\text{ Hz}$ ($k=25$, $s=5$, max-pool 2)\newline Identity embeddings on keys ($\gamma_0 = 0.5$)} \\
\midrule
\TCell{II}{3}{0}{\strong{Ninapro DB5} \cite{ref3}\newline Intact-Limb Wearable Armband} & \TCell{II}{3}{1}{Dual Thalmic Myo Armbands\newline  $f_s = 200\text{ Hz}$ \newline (Pre-conditioned sEMG)} & \TCell{II}{3}{2}{$16\text{-EMG} + 3\text{-ACC}$ \newline  $T = 40$, stride $10$ \newline Train: rep $\{1,3,4,6\}$ \newline Test: rep $\{2,5\}$} & \TCell{II}{3}{3}{DB5-All10 ($n = 10$: s1–s10)\newline DB5-Holdout5 ($n = 5$: s6–s10) $^{\text{b}}$ \newline DB5-Dev5 ($n = 5$: s1–s5) $^{\text{b}}$} & \TCell{II}{3}{4}{B17 (17) \& ABC52 (52)\newline B17 (17) \& ABC52 (52)\newline B17 (17) \& ABC52 (52)} & \TCell{II}{3}{5}{Front end for $200\text{ Hz}$ (strides $1, 2, 2$; no $10\times$ down-sampling)\newline Identity embeddings on queries only ($\gamma_0 = 0.1$)} \\
\bottomrule
\end{tabular}
\par\vspace{3pt}
\begin{minipage}{\textwidth}\footnotesize
\TNote{II}{0}{$^{\text{a}}$ All setups share latent dimension $D = 64$, $H = 4$ attention heads, $\lambda_{\text{gate}} = 0.5$, $\lambda_{\text{impute}} = 0.2$, AdamW ($\text{lr} = 10^{-3}$, weight decay $10^{-4}$), batch size $256$, and 40 training epochs across 3 seeds (20261025–20261027).}\par
\TNote{II}{1}{$^{\text{b}}$ Nested subcohorts within the corresponding full dataset cohort (DB3-All9, DB2-All10, or DB5-All10); reported to document the development/confirmation split, not as additional independent subjects.}\par
\end{minipage}
\endgroup
\end{table*}

\subsection{Synthetic Corruption Operators and Training Protocol}

Perturbations were applied in the standardized signal domain across $K$ randomly selected sEMG channels, with $m_c = 0$, while $m_{c'} = 1$ for the remaining $C_e - K$ channels \cite{ref9}, \cite{ref10}, \cite{ref11}. For compound motion-drift and broadband burst artifacts, termed Burst, each selected channel $c$ is perturbed over a contiguous sub-window $[t_{0,c},\, t_{0,c} + L_c)$ with $t_{0,c} \in \{0, \dots, \lfloor T/2 \rfloor - 1\}$ and $L_c \in \{\lfloor T/4 \rfloor, \dots, \lfloor T/2 \rfloor\}$ by a low-frequency baseline-wander sinusoid with $f_c \sim \mathcal{U}(1, 10)\text{ Hz}$ and $\phi_c \sim \mathcal{U}(0, 2\pi)$ superimposed with broadband Gaussian noise $\xi_{c,t} \sim \mathcal{N}(0, 1)$ \cite{ref10}, \cite{ref11}, scaled to a target window-level signal-to-noise ratio $\text{SNR}_{\text{dB}}$ using the whole-window root-mean-square energy $\text{RMS}(X_c) = \max\big(\sqrt{\frac{1}{T}\sum_{t=1}^{T} X_{c,t}^2},\, 10^{-12}\big)$ and the corresponding whole-window perturbation energy $\text{RMS}(\tilde{\eta}_c)$ computed over all $T$ samples:

\begin{equation}
\tilde{\eta}_{c,t} = \big(\sin(2\pi f_c t / f_s + \phi_c) + 0.5\,\xi_{c,t}\big) \cdot \mathbb{I}\big(t_{0,c} \le t < t_{0,c} + L_c\big)
\label{eq:14}
\end{equation}

\begin{equation}
X_{c,t}^{\text{corrupt}} = X_{c,t} + \tilde{\eta}_{c,t} \cdot \frac{\text{RMS}(X_c)}{\text{RMS}(\tilde{\eta}_c)} \cdot 10^{-\text{SNR}_{\text{dB}} / 20}
\label{eq:15}
\end{equation}

For transient electrode lift-off shocks, termed Liftoff, loss of common-mode rejection is modeled by injecting a high-amplitude $50\text{ Hz}$ powerline sinusoid coupled with a random DC step offset:

\begin{equation}
\begin{gathered}\eta_{c,t}^{\text{lift}} = A_c \sin(2\pi \cdot 50\, t / f_s + \phi_c) + A_c \delta_c,\\ A_c \sim \mathcal{U}(5, 10), \quad \delta_c \sim \mathcal{U}(-1, 1)\end{gathered}
\label{eq:16}
\end{equation}

where $X_{c,t}^{\text{corrupt}} = X_{c,t} + \eta_{c,t}^{\text{lift}}$ is applied directly in standardized channel units during training and main-table evaluations under Liftoff-25\% and Liftoff-50\%, and is rescaled via (15) during controlled $\text{SNR}_{\text{dB}}$ sweeps. Following the zero-masking convention in incomplete multimodal fusion \cite{ref13}, channel dropout, termed Dropout, serves as a digital proxy for signal loss by setting selected standardized channels to $X_{c,t}^{\text{corrupt}} = 0$, whereas whole-modality dropout zeros the entire sEMG array in EMG-loss or the inertial stream in ACC-loss.

During training of all corruption-augmented models, each window is assigned with equal probability of $1/6$ to clean, Burst, Liftoff, Dropout, EMG-loss, or ACC-loss, so partial channel corruption occurs with probability $1/2$ by drawing $K \sim \mathcal{U}\{1, \dots, C_e/2\}$ and Burst $\text{SNR}_{\text{dB}} \sim \mathcal{U}\{0, -5, -10, -15\}\text{ dB}$. At test time, models are evaluated without retraining on clean windows; under $25\%$ and $50\%$ partial corruption with $K = C_e/4$ and $K = C_e/2$, denoted Burst-25\%/50\% at $-10\text{ dB}$, Dropout-25\%/50\%, and Liftoff-25\%/50\%, with Dropout-25\% omitted on DB5; and under EMG-loss and ACC-loss. We also sweep $K \in \{0, 2, \dots, C_e\}$ and $\text{SNR}_{\text{dB}} \in \{+5, 0, -5, -10, -15\}\text{ dB}$ at $K = C_e/2$ across all seven models on DB3 and DB2 and the five core models on DB5, averaging three matched deterministic corruption draws per window so that all models receive identical corrupted inputs.

\subsection{Compared Baselines, Ablation Variants, and Implementation Details}

We first define four reference models. ACC-Only uses only the inertial branch, trained on clean data. Vanilla-Clean is a conventional 1-D CNN that mixes all sEMG channels in its first layer using one filter group, applies feature-wise layer normalization after global temporal pooling, concatenates the sEMG and inertial features, and is trained on clean data only. Vanilla-Aug has the same architecture as Vanilla-Clean but is trained under the six-condition corruption distribution described above. Gate-Only combines the channel-independent encoder, which uses $C_e$ filter groups, with the multi-statistic soft gate configured with $\lambda_{\text{gate}} = 0.5$ and pools $G_{\text{emg}} = \boldsymbol{\alpha} \odot Z_{\text{emg}}$ directly, without imputation; it isolates the effect of gating and serves as the main control for QA-CMIF.

Relative to QA-CMIF, we evaluated three single-factor ablations, where w/o Bypass sets $H_{\text{emg}} = \tilde{Z}_{\text{imp}}$ in (10), w/o Align Loss sets $\lambda_{\text{impute}} = 0$, and w/o Stop-Grad removes stop-gradient detachment on $\bar{\boldsymbol{\alpha}}$. For external comparison, we adapted lightweight raw-window versions of HyFusion \cite{ref20} and AiFusion \cite{ref13}, termed HyFusion-Lite and AiFusion-Lite, using the same 3-ACC input, sampling-rate-specific temporal stem, and feature dimension $D = 64$. HyFusion-Lite retains its three parallel sEMG, ACC, and joint residual-attention branches trained with multi-branch cross-entropy and center-repulsion island loss \cite{ref20}, while AiFusion-Lite retains its parallel multi-scale residual stems and three-layer four-head interactive Transformer encoder trained with multi-branch cross-entropy, supervised contrastive loss, and online self-distillation Kullback--Leibler loss \cite{ref13}. All corruption-augmented baselines, namely Vanilla-Aug, Gate-Only, HyFusion-Lite, and AiFusion-Lite, were trained under the identical paired clean-plus-corrupted objective $\mathcal{L}_{\text{clean}} + \mathcal{L}_{\text{corrupt}}$ in PyTorch using AdamW \cite{ref35} with learning rate $10^{-3}$, weight decay $10^{-4}$, cosine annealing, gradient clipping at $5.0$, and batch size $256$ for 40 epochs across seeds 20261025--20261027, evaluating final-epoch weights directly.

\subsection{Hierarchical Paired Statistical Analysis}

All statistical comparisons treat the participant as the independent experimental unit of sample size $n$ after averaging the three corruption draws and three training seeds within each subject. Paired differences $\Delta_i = y_{i,\text{QA-CMIF}} - y_{i,\text{baseline}}$ in percentage points, denoted pp, were evaluated using two-sided paired Student's $t$-tests and exact Wilcoxon signed-rank tests alongside participant win/loss counts. To control the family-wise error rate \cite{ref36}, we define a primary comparison family per cohort and task comprising Burst-50\%, Dropout-50\%, and Liftoff-50\% against the five non-ablation baselines on B17 and BC40 for 15 comparisons and against the three core baselines on ABC52 for 9 comparisons. Unless noted otherwise, $p_{\text{Holm}}$ in the main text denotes two-sided paired-$t$ Holm-adjusted $p$-values within this primary family, whereas Wilcoxon tests, clean and $25\%$ corruption conditions, ACC-Only contrasts, and ablations are reported with unadjusted two-sided paired $p$-values, denoted $p_{\text{raw}}$. Complete 95\% confidence intervals, Cohen's $d_z$, and full-table Holm adjustments across all contrasts are provided in Supplementary Table S3.

\begin{figure*}[!t]
\centering
\includegraphics[width=.78\textwidth,trim=0bp 8.5bp 0bp 0bp,clip]{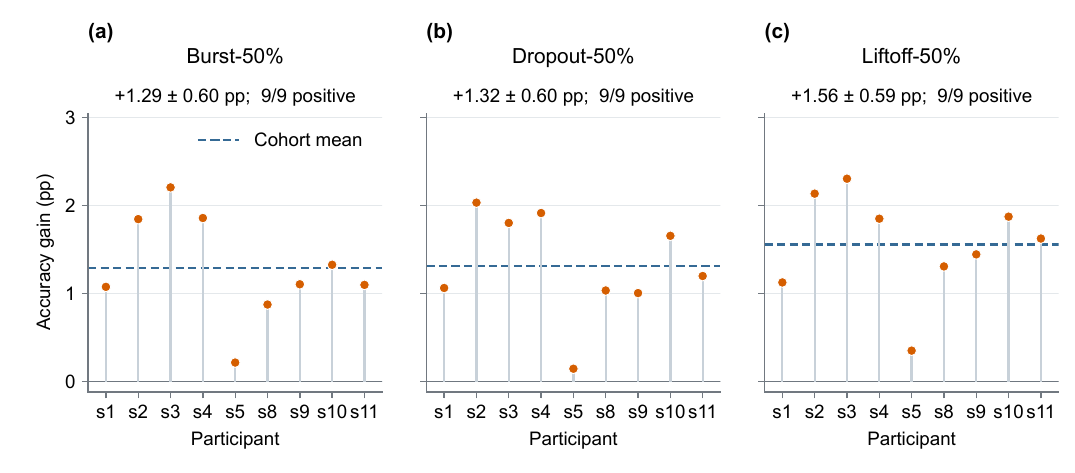}
\caption{Participant-level accuracy gains of QA-CMIF over Gate-Only on DB3-All9 for the 17-class B17 task under 50\% channel corruption. Orange markers show individual participant gains computed as the difference in three-seed mean accuracy between QA-CMIF and Gate-Only, blue dashed lines indicate the cohort mean, and panel headers report the cohort mean $\pm$ SD alongside the count of participants with positive gains.}
\label{fig:2}
\end{figure*}
\section{Results}

\subsection{Primary Evaluation on Transradial Amputees in Ninapro DB3 at $2000\text{ Hz}$}

We first test whether cross-modal latent imputation improves fault-tolerant myoelectric decoding on clinical residual limbs without degrading uncorrupted accuracy.
Table III compares QA-CMIF against inertial-only, ungated, gate-only, and adapted multimodal fusion baselines on the transradial amputee cohorts of Ninapro DB3 at $2000\text{ Hz}$ with 12-EMG and 3-ACC inputs across the 17-class B17 and 40-class BC40 benchmarks \cite{ref6}, where all classification accuracies and Macro-F1 scores are reported in \% and paired model differences are reported in percentage points, denoted pp.
On the nine-amputee DB3-All9 cohort on B17, QA-CMIF achieves $74.85 \pm 12.63\%$ clean accuracy and $74.00 \pm 12.05\%$ Macro-F1, matching the Gate-Only control at $74.93\%$ with $p = 0.710$ while surpassing Vanilla-Aug at $68.13\%$ by $+6.72\text{ pp}$ with $p = 4.47 \times 10^{-5}$, Vanilla-Clean at $65.58\%$ by $+9.27\text{ pp}$, and ACC-Only at $58.67\%$ by $+16.19\text{ pp}$.

Under $50\%$ corruption with $K = 6/12$ on DB3-All9, QA-CMIF attains $69.97\%$ to $70.04\%$ accuracy across Burst, Dropout, and Liftoff, outperforming Gate-Only at $68.48\%$ to $68.67\%$ by $+1.29\text{ to }{+}1.56\text{ pp}$ with $9/0$ subject wins in every regime, paired-$t$ $p_{\text{Holm}} \le 0.00124$, unadjusted Wilcoxon $p = 0.00391$, and 15-test Wilcoxon $p_{\text{Holm}} = 0.0586$, as shown in Fig.~\ref{fig:2}. Macro-F1 gains over Gate-Only span $+1.44\text{ to }{+}1.72\text{ pp}$ with $p_{\text{raw}} \le 1.56 \times 10^{-4}$, and $25\%$-corruption accuracy gains span $+0.51\text{ to }{+}0.61\text{ pp}$ with $8/9$ wins and $p_{\text{raw}} \le 0.0215$ in Supplementary Table S2. Relative to Vanilla-Aug, $50\%$-corruption gains span $+6.26\text{ to }{+}10.71\text{ pp}$ with $p_{\text{Holm}} \le 2.47 \times 10^{-4}$, whereas under complete modality loss QA-CMIF achieves $58.00\%$ under EMG-loss compared with $58.67\%$ for ACC-Only and $44.98\%$ under ACC-loss compared with $45.40\%$ for Gate-Only.

Across the pre-specified subcohorts in Supplementary Table S2, QA-CMIF achieves $6/6$ subject wins over Gate-Only on DB3-Core6 under B17 $50\%$ corruption with gains of $+1.17\text{ to }{+}1.45\text{ pp}$ and $p_{\text{Holm}} \le 0.0182$, and $3/3$ wins on DB3-Holdout3 with gains of $+1.51\text{ to }{+}1.77\text{ pp}$ and $p_{\text{raw}} \le 0.0463$. On the 40-class BC40 benchmark on DB3-Core6 in Table III, QA-CMIF attains $69.68 \pm 5.86\%$ clean accuracy and $62.30\%$ to $62.46\%$ accuracy under $50\%$ corruption, where its gain over Gate-Only grows by $1.85\text{--}2.22\times$ to $+2.59\text{ to }{+}2.69\text{ pp}$ with $6/6$ wins across all three modes and $p_{\text{Holm}} \le 1.22 \times 10^{-4}$.

Relative to adapted external baselines under $50\%$ corruption, QA-CMIF surpasses AiFusion-Lite \cite{ref13} by $+4.91\text{ to }{+}11.06\text{ pp}$ on B17 and $+7.50\text{ to }{+}16.60\text{ pp}$ on BC40, with respective $p_{\text{Holm}}$ upper bounds of $0.00372$ and $0.0140$. Against the $5.23\times$ heavier HyFusion-Lite \cite{ref20}, clean gains span $+1.96\text{ to }{+}4.45\text{ pp}$ and Burst-50\% margins span $+0.78\text{ to }{+}1.29\text{ pp}$, while Dropout-50\% and Liftoff-50\% gains reach $+5.05\text{ to }{+}9.13\text{ pp}$ and $+9.00\text{ to }{+}11.17\text{ pp}$, with $p_{\text{Holm}} \le 0.00865$ and $0.0140$, respectively.
Across both 17-class and 40-class clinical amputee benchmarks, QA-CMIF preserves uncorrupted decoding accuracy while consistently outperforming gate-only attenuation and corruption-augmented fusion baselines under partial sEMG channel degradation.

\begin{table*}[!t]
\centering\begingroup
\fontsize{7.5}{8.5}\selectfont
\setlength{\tabcolsep}{3.5pt}
\renewcommand{\arraystretch}{0.95}
\caption{PRIMARY CLINICAL BENCHMARK PERFORMANCE ON TRANSRADIAL AMPUTEES (NINAPRO DB3, $2000\text{ Hz}$, $12\text{-EMG} + 3\text{-ACC}$, MEAN $\pm$ SD \% ACROSS SUBJECTS)}
\label{tab:III}
\begin{tabular}{@{}lccccccc@{}}
\toprule
Model & \shortstack{clean Acc (F1)\\(\%)} & \shortstack{Burst-50\%\\(\%)} & \shortstack{Dropout-50\%\\(\%)} & \shortstack{Liftoff-50\%\\(\%)} & \shortstack{EMG-loss\\(\%)} & \shortstack{ACC-loss\\(\%)} & \shortstack{Liftoff gain (pp)\\/ paired $p$ $^{\mathrm a}$}\\\midrule
\midrule\multicolumn{8}{@{}l}{\strong{DB3-All9}  ($n = 9$ amputees,  B17 17 classes)}\\
ACC-Only & 58.67 $\pm$ 15.01 (56.20) & 58.67 $\pm$ 15.01 & 58.67 $\pm$ 15.01 & 58.67 $\pm$ 15.01 & \strong{58.67 $\pm$ 15.01} & 5.38 $\pm$ 2.03 & +11.38 ($9.0 \times 10^{-5}$)\\
Vanilla-Clean & 65.58 $\pm$ 12.31 (64.44) & 37.72 $\pm$ 11.73 & 50.27 $\pm$ 12.77 & 32.01 $\pm$ 10.88 & 31.01 $\pm$ 11.13 & 20.24 $\pm$ 3.62 & +38.03 ($7.5 \times 10^{-5}$)\\
Vanilla-Aug & 68.13 $\pm$ 12.81 (67.12) & 63.70 $\pm$ 14.53 & 60.49 $\pm$ 13.99 & 59.33 $\pm$ 15.16 & 58.53 $\pm$ 14.33 & 40.63 $\pm$ 10.26 & +10.71 ($1.1 \times 10^{-4}$)\\
HyFusion-Lite \cite{ref20} & 72.89 $\pm$ 13.04 (71.97) & 69.19 $\pm$ 14.57 & 64.92 $\pm$ 14.65 & 61.04 $\pm$ 16.46 & 58.10 $\pm$ 15.21 & \strong{47.00 $\pm$ 9.93} & +9.00 ($0.00252$)\\
AiFusion-Lite \cite{ref13} & 68.72 $\pm$ 14.39 (67.70) & 65.06 $\pm$ 15.59 & 61.24 $\pm$ 15.33 & 58.99 $\pm$ 16.65 & 58.50 $\pm$ 16.17 & 44.69 $\pm$ 10.86 & +11.06 ($0.00180$)\\
Gate-Only & \strong{74.93 $\pm$ 12.39} (\strong{74.01}) & 68.67 $\pm$ 13.81 & 68.65 $\pm$ 13.77 & 68.48 $\pm$ 13.82 & 57.89 $\pm$ 14.94 & 45.40 $\pm$ 8.48 & Ref.\\
\strong{QA-CMIF} & 74.85 $\pm$ 12.63 (74.00) & \strong{69.97 $\pm$ 13.48} & \strong{69.97 $\pm$ 13.48} & \strong{70.04 $\pm$ 13.49} & 58.00 $\pm$ 14.38 & 44.98 $\pm$ 8.69 & \strong{+1.56 $\pm$ 0.59} (9/9, $4.3 \times 10^{-4}$)\\
\midrule\multicolumn{8}{@{}l}{\strong{DB3-Core6}  ($n = 6$ amputees,  BC40 40 classes)}\\
ACC-Only & 44.70 $\pm$ 8.62 (43.50) & 44.70 $\pm$ 8.62 & 44.70 $\pm$ 8.62 & 44.70 $\pm$ 8.62 & \strong{44.70 $\pm$ 8.62} & 2.70 $\pm$ 0.60 & +17.75 ($1.7 \times 10^{-4}$)\\
Vanilla-Clean & 60.59 $\pm$ 7.97 (60.32) & 25.68 $\pm$ 5.52 & 39.29 $\pm$ 8.62 & 20.83 $\pm$ 4.14 & 18.39 $\pm$ 6.87 & 13.66 $\pm$ 3.90 & +41.62 ($2.2 \times 10^{-4}$)\\
Vanilla-Aug & 62.81 $\pm$ 8.73 (62.57) & 56.35 $\pm$ 10.03 & 51.21 $\pm$ 8.62 & 47.91 $\pm$ 10.26 & 43.85 $\pm$ 7.56 & 32.08 $\pm$ 8.06 & +14.54 ($0.00533$)\\
HyFusion-Lite \cite{ref20} & 65.23 $\pm$ 7.74 (65.38) & 61.11 $\pm$ 8.93 & 53.17 $\pm$ 9.43 & 51.28 $\pm$ 10.92 & 42.38 $\pm$ 8.72 & 36.79 $\pm$ 7.73 & +11.17 ($0.01397$)\\
AiFusion-Lite \cite{ref13} & 59.61 $\pm$ 8.39 (59.94) & 54.90 $\pm$ 9.32 & 48.24 $\pm$ 9.86 & 45.85 $\pm$ 10.67 & 43.84 $\pm$ 9.04 & 33.75 $\pm$ 7.45 & +16.60 ($0.00793$)\\
Gate-Only & 69.32 $\pm$ 5.93 (69.00) & 59.80 $\pm$ 6.55 & 59.71 $\pm$ 6.34 & 59.76 $\pm$ 6.43 & 43.65 $\pm$ 7.08 & \strong{39.98 $\pm$ 6.54} & Ref.\\
\strong{QA-CMIF} & \strong{69.68 $\pm$ 5.86} (\strong{69.30}) & \strong{62.40 $\pm$ 6.32} & \strong{62.30 $\pm$ 6.36} & \strong{62.46 $\pm$ 6.40} & 44.01 $\pm$ 7.28 & 39.62 $\pm$ 6.50 & \strong{+2.69 $\pm$ 0.25} (6/6, $1.9 \times 10^{-5}$)\\
\bottomrule\end{tabular}
\par\vspace{2pt}
\begin{minipage}{\textwidth}\scriptsize
$^{\text{a}}$ For each baseline row, $\Delta = \text{QA-CMIF} - \text{row model}$ under Liftoff-50\%, with 15-comparison primary-family paired-$t$ $p_{\text{Holm}}$ for the five primary baselines and unadjusted paired-$t$ $p_{\text{raw}}$ for ACC-Only; the QA-CMIF row highlights the net imputation gain $\text{QA-CMIF} - \text{Gate-Only}$, with subject wins/total participants and $p_{\text{Holm}}$. Complete 9-condition results including $25\%$ corruption and nested subcohorts are provided in Supplementary Table S2. Boldface indicates the highest accuracy in each column within each cohort, and clean-condition Macro-F1 is reported in parentheses next to clean accuracy. All accuracy and Macro-F1 values are in \%, and all paired gains $\Delta$ are in percentage points (pp).
\end{minipage}\endgroup
\end{table*}

\begin{figure}[!t]
\centering
\includegraphics[width=.97\columnwidth,trim=0bp 29.5bp 0bp 0bp,clip]{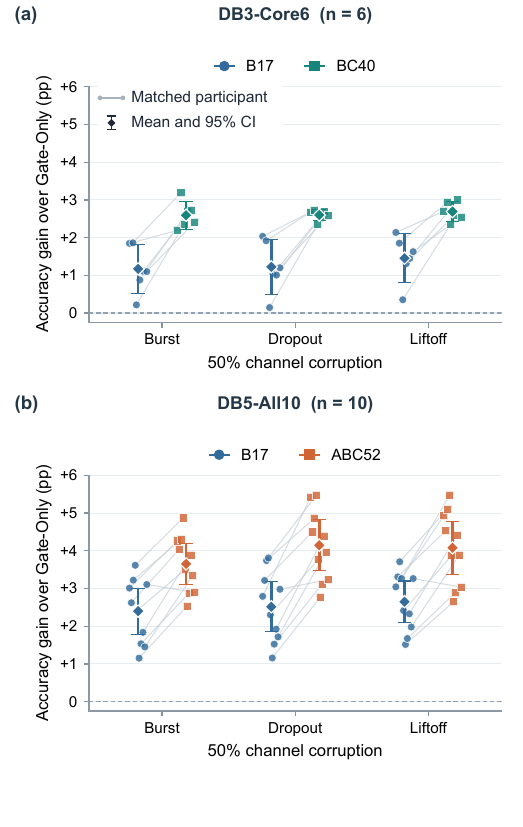}
\caption{Paired participant-level accuracy gains of QA-CMIF over Gate-Only under 50\% channel corruption comparing B17 with BC40 on DB3-Core6 and B17 with ABC52 on DB5-All10. Markers connected by thin gray lines represent matched participant gains computed from three-seed mean accuracies, and dark markers with vertical error bars indicate the cohort mean and two-sided 95\% Student's t confidence interval across participants.}
\label{fig:3}
\end{figure}
\subsection{Evaluation Across Acquisition Setups on Intact Cohorts from Ninapro DB2 and DB5}

Assessing whether the robustness gains of QA-CMIF generalize beyond clinical amputees requires validation across intact-limb populations and distinct sensor hardware.
Table IV evaluates all models on two 10-subject intact cohorts, comprising $2000\text{ Hz}$ Delsys Trigno medical arrays with 12-EMG and 3-ACC inputs in DB2-All10 on B17 \cite{ref6} and $200\text{ Hz}$ consumer dual-Myo armbands with 16-EMG and 3-ACC inputs in DB5-All10 across B17 and the 52-class ABC52 repertoire \cite{ref3}.
On DB2-All10 B17, QA-CMIF achieves $87.72 \pm 3.66\%$ clean accuracy, matching Gate-Only at $87.58\%$ with $p = 0.399$ while exceeding Vanilla-Aug at $82.63\%$ and AiFusion-Lite at $83.03\%$. Under $50\%$ corruption with $K = 6/12$, QA-CMIF attains $81.03\%$ to $81.28\%$ accuracy across Burst, Dropout, and Liftoff, outperforming Gate-Only by $+1.67\text{ to }{+}1.74\text{ pp}$ with $10/0$ subject wins, paired-$t$ $p_{\text{Holm}} \le 8.05 \times 10^{-5}$, and Wilcoxon $p_{\text{Holm}} = 0.0293$. The five-subject confirmation split DB2-Holdout5 likewise achieves $5/5$ wins with gains of $+1.56\text{ to }{+}1.72\text{ pp}$ and $p_{\text{Holm}} = 0.0320$ in Supplementary Table S2.

Under DB2 Burst-50\%, QA-CMIF at $81.28\%$ and the $5.23\times$ heavier HyFusion-Lite at $82.10\%$ show no statistically significant difference, with $\Delta = -0.81 \pm 2.70\text{ pp}$ and $p_{\text{raw}} = 0.365$. Under Dropout-50\% and Liftoff-50\%, however, HyFusion-Lite drops to $75.18\%$ and $68.71\%$, and QA-CMIF surpasses it in all $10/10$ subjects by $+5.85\text{ pp}$ and $+12.56\text{ pp}$, respectively, with $p_{\text{Holm}} \le 2.26 \times 10^{-4}$.

On DB5-All10 B17, QA-CMIF attains $83.08 \pm 4.02\%$ clean accuracy and $78.01\%$ to $78.18\%$ accuracy under $50\%$ corruption with $K = 8/16$, outperforming Gate-Only by $+2.40\text{ to }{+}2.65\text{ pp}$ with $10/0$ wins and $p_{\text{Holm}} \le 5.26 \times 10^{-5}$. Gains over HyFusion-Lite and AiFusion-Lite span $+1.62\text{ to }{+}14.12\text{ pp}$ with $p_{\text{Holm}} \le 0.0441$ and $+4.97\text{ to }{+}12.43\text{ pp}$ with $p_{\text{Holm}} \le 1.08 \times 10^{-4}$, respectively. On the five-subject confirmation split DB5-Holdout5, gains over Gate-Only of $+1.82\text{ to }{+}2.15\text{ pp}$ achieve $5/5$ wins with $p_{\text{raw}} \le 0.00595$ and $p_{\text{Holm}} \le 0.0214$ in Supplementary Tables S2 and S3. On the 52-class ABC52 benchmark, QA-CMIF achieves $80.27 \pm 4.37\%$ clean accuracy and $73.33\%$ to $73.76\%$ accuracy under $50\%$ corruption, expanding its gains over Gate-Only to $+3.65\text{ to }{+}4.15\text{ pp}$ with $10/10$ wins and $p_{\text{Holm}} \le 7.13 \times 10^{-7}$, alongside $+6.76\text{ to }{+}19.85\text{ pp}$ gains over Vanilla-Aug.
QA-CMIF thus generalizes across both $2000\text{ Hz}$ medical arrays and $200\text{ Hz}$ consumer armbands on intact subjects, maintaining clean-condition parity with Gate-Only while improving accuracy under partial sEMG corruption across 17-class and 52-class gesture sets.

\begin{table*}[!t]
\centering\begingroup
\fontsize{7.5}{8.5}\selectfont
\setlength{\tabcolsep}{3.5pt}
\renewcommand{\arraystretch}{0.95}
\caption{CROSS-SETUP BENCHMARK PERFORMANCE ON INTACT COHORTS (NINAPRO DB2 AT $2000\text{ Hz}$ AND NINAPRO DB5 AT $200\text{ Hz}$, MEAN $\pm$ SD \% ACROSS SUBJECTS)}
\label{tab:IV}
\begin{tabular}{@{}lccccccc@{}}
\toprule
Model & \shortstack{clean Acc (F1)\\(\%)} & \shortstack{Burst-50\%\\(\%)} & \shortstack{Dropout-50\%\\(\%)} & \shortstack{Liftoff-50\%\\(\%)} & \shortstack{EMG-loss\\(\%)} & \shortstack{ACC-loss\\(\%)} & \shortstack{Liftoff gain (pp)\\/ paired $p$ $^{\mathrm a}$}\\\midrule
\midrule\multicolumn{8}{@{}l}{\strong{DB2-All10}  ($2000\text{ Hz}$, 12-EMG,   $n = 10$, B17 17 cl.)}\\
ACC-Only & 65.19 $\pm$ 6.51 (63.81) & 65.19 $\pm$ 6.51 & 65.19 $\pm$ 6.51 & 65.19 $\pm$ 6.51 & 65.19 $\pm$ 6.51 & 6.80 $\pm$ 1.76 & +16.08 ($1.2 \times 10^{-6}$)\\
Vanilla-Clean & 80.92 $\pm$ 5.83 (80.72) & 43.91 $\pm$ 6.30 & 58.41 $\pm$ 7.33 & 29.93 $\pm$ 4.54 & 28.76 $\pm$ 10.31 & 30.74 $\pm$ 10.69 & +51.34 ($1.7 \times 10^{-9}$)\\
Vanilla-Aug & 82.63 $\pm$ 6.55 (82.64) & 76.67 $\pm$ 8.11 & 71.03 $\pm$ 7.05 & 66.15 $\pm$ 6.87 & 64.56 $\pm$ 6.94 & 59.83 $\pm$ 9.45 & +15.12 ($5.5 \times 10^{-6}$)\\
HyFusion-Lite \cite{ref20} & 86.47 $\pm$ 5.44 (86.39) & \strong{82.10 $\pm$ 7.05} & 75.18 $\pm$ 6.76 & 68.71 $\pm$ 7.61 & 63.56 $\pm$ 7.75 & \strong{68.84 $\pm$ 9.14} & +12.56 ($8.2 \times 10^{-6}$)\\
AiFusion-Lite \cite{ref13} & 83.03 $\pm$ 6.51 (83.13) & 77.71 $\pm$ 7.51 & 71.16 $\pm$ 6.61 & 66.53 $\pm$ 6.65 & \strong{65.32 $\pm$ 6.53} & 66.42 $\pm$ 9.48 & +14.73 ($1.7 \times 10^{-6}$)\\
Gate-Only & 87.58 $\pm$ 3.75 (87.43) & 79.61 $\pm$ 5.20 & 79.28 $\pm$ 5.31 & 79.57 $\pm$ 5.33 & 63.62 $\pm$ 6.94 & 66.68 $\pm$ 6.59 & Ref.\\
\strong{QA-CMIF} & \strong{87.72 $\pm$ 3.66} (\strong{87.56}) & 81.28 $\pm$ 4.72 & \strong{81.03 $\pm$ 4.78} & \strong{81.26 $\pm$ 4.86} & 64.08 $\pm$ 7.39 & 66.06 $\pm$ 6.09 & \strong{+1.69 $\pm$ 0.61} (10/10, $7.9 \times 10^{-5}$)\\
\midrule\multicolumn{8}{@{}l}{\strong{DB5-All10}  ($200\text{ Hz}$, 16-EMG,   $n = 10$, B17 17 cl.)}\\
ACC-Only & 64.58 $\pm$ 6.84 (63.52) & 64.58 $\pm$ 6.84 & 64.58 $\pm$ 6.84 & 64.58 $\pm$ 6.84 & 64.58 $\pm$ 6.84 & 5.89 $\pm$ 0.87 & +13.54 ($3.6 \times 10^{-7}$)\\
Vanilla-Clean & 78.53 $\pm$ 4.04 (78.41) & 48.29 $\pm$ 5.25 & 61.41 $\pm$ 4.27 & 34.64 $\pm$ 5.78 & 37.10 $\pm$ 9.83 & 28.01 $\pm$ 5.55 & +43.47 ($7.3 \times 10^{-9}$)\\
Vanilla-Aug & 80.49 $\pm$ 4.30 (80.44) & 72.26 $\pm$ 5.74 & 68.71 $\pm$ 4.67 & 63.63 $\pm$ 6.96 & 63.64 $\pm$ 7.04 & 51.60 $\pm$ 4.82 & +14.48 ($5.1 \times 10^{-6}$)\\
HyFusion-Lite \cite{ref20} & \strong{83.81 $\pm$ 3.33} (\strong{83.95}) & 76.57 $\pm$ 5.57 & 72.25 $\pm$ 4.77 & 63.99 $\pm$ 7.64 & 64.30 $\pm$ 7.04 & \strong{61.31 $\pm$ 4.34} & +14.12 ($1.5 \times 10^{-5}$)\\
AiFusion-Lite \cite{ref13} & 78.42 $\pm$ 4.00 (78.51) & 73.21 $\pm$ 5.80 & 69.63 $\pm$ 5.48 & 65.68 $\pm$ 7.31 & \strong{65.37 $\pm$ 7.24} & 53.57 $\pm$ 5.46 & +12.43 ($1.7 \times 10^{-5}$)\\
Gate-Only & 83.04 $\pm$ 4.18 (82.96) & 75.78 $\pm$ 4.97 & 75.50 $\pm$ 4.86 & 75.46 $\pm$ 4.83 & 62.27 $\pm$ 7.18 & 51.57 $\pm$ 5.18 & Ref.\\
\strong{QA-CMIF} & 83.08 $\pm$ 4.02 (82.99) & \strong{78.18 $\pm$ 4.42} & \strong{78.01 $\pm$ 4.34} & \strong{78.11 $\pm$ 4.37} & 63.89 $\pm$ 7.08 & 51.07 $\pm$ 5.27 & \strong{+2.65 $\pm$ 0.77} (10/10, $1.4 \times 10^{-5}$)\\
\midrule\multicolumn{8}{@{}l}{\strong{DB5-All10}  ($200\text{ Hz}$, 16-EMG,   $n = 10$, ABC52 52 cl.)}\\
ACC-Only & 54.71 $\pm$ 6.85 (52.94) & 54.71 $\pm$ 6.85 & 54.71 $\pm$ 6.85 & 54.71 $\pm$ 6.85 & \strong{54.71 $\pm$ 6.85} & 1.89 $\pm$ 0.24 & +19.05 ($1.3 \times 10^{-8}$)\\
Vanilla-Clean & 73.88 $\pm$ 4.08 (73.39) & 35.13 $\pm$ 4.21 & 51.02 $\pm$ 3.86 & 23.17 $\pm$ 3.39 & 23.24 $\pm$ 4.66 & 16.07 $\pm$ 3.91 & +50.59 ($7.6 \times 10^{-12}$)\\
Vanilla-Aug & 76.87 $\pm$ 3.88 (76.47) & 66.58 $\pm$ 4.62 & 61.47 $\pm$ 4.17 & 53.91 $\pm$ 6.33 & 53.52 $\pm$ 6.58 & 44.31 $\pm$ 4.10 & +19.85 ($4.5 \times 10^{-8}$)\\
Gate-Only & 80.01 $\pm$ 4.40 (79.63) & 69.68 $\pm$ 4.75 & 69.49 $\pm$ 4.92 & 69.68 $\pm$ 4.86 & 52.58 $\pm$ 6.71 & 47.37 $\pm$ 4.61 & Ref.\\
\strong{QA-CMIF} & \strong{80.27 $\pm$ 4.37} (\strong{79.88}) & \strong{73.33 $\pm$ 4.54} & \strong{73.64 $\pm$ 4.61} & \strong{73.76 $\pm$ 4.61} & 53.24 $\pm$ 6.96 & \strong{47.74 $\pm$ 4.64} & \strong{+4.08 $\pm$ 0.98} (10/10, $7.1 \times 10^{-7}$)\\
\bottomrule\end{tabular}
\par\vspace{2pt}
\begin{minipage}{\textwidth}\scriptsize
$^{\text{a}}$ For each baseline row, $\Delta = \text{QA-CMIF} - \text{row model}$ under Liftoff-50\%, with 15/9-comparison primary-family paired-$t$ $p_{\text{Holm}}$ for the primary baselines and unadjusted paired-$t$ $p_{\text{raw}}$ for ACC-Only; the QA-CMIF row highlights $\text{QA-CMIF} - \text{Gate-Only}$. Complete results for $25\%$ corruption through Burst-25\%, Dropout-25\% on DB2, and Liftoff-25\%, and for the 5-subject splits DB2-Holdout5 and DB5-Holdout5, are provided in Supplementary Table S2. Boldface indicates the highest accuracy in each column within each cohort, and clean-condition Macro-F1 is reported in parentheses next to clean accuracy. All accuracy and Macro-F1 values are in \%, and all paired gains $\Delta$ are in percentage points (pp).
\end{minipage}\endgroup
\end{table*}

\subsection{Exploratory Analysis of Imputation Gain Across Gesture Set Cardinalities}

Comparing the 17-class B17 results in Tables III and IV with those on BC40 and ABC52 indicates that the accuracy benefit of latent imputation over gate-only attenuation grows on larger gesture repertoires.
To visualize this scaling behavior at the participant level, Fig.~\ref{fig:3} plots side-by-side paired scatter comparisons of the QA-CMIF accuracy gain over Gate-Only under $50\%$ channel corruption, contrasting the 17-class B17 task against the 40-class BC40 task on DB3-Core6 in the top panel and against the 52-class ABC52 task on DB5-All10 in the bottom panel, with full $25\%$ and $50\%$ paired statistics reported in Supplementary Table S5. Because expanding from B17 to BC40 or ABC52 jointly alters class cardinality, movement composition, and training sample volume without a subsampled 17-class control, this comparison is treated as an exploratory within-subject analysis.

In the top panel of Fig.~\ref{fig:3}, every paired participant line on the six-amputee DB3-Core6 cohort slopes upward from B17 to BC40 across all three $50\%$ corruption modes, reflecting higher imputation gains on BC40 in all $6/6$ amputees with exact Wilcoxon $p = 0.0313$. Specifically, the cohort-mean accuracy gain over Gate-Only rises by $1.85\text{--}2.22\times$ from $+1.17\text{--}1.45\text{ pp}$ on B17 to $+2.59\text{--}2.69\text{ pp}$ on BC40, representing net increases of $+1.43\text{ pp}$ under Burst, $+1.37\text{ pp}$ under Dropout, and $+1.24\text{ pp}$ under Liftoff with paired-$t$ $p \le 0.00237$. Under $25\%$ corruption in Supplementary Table S5, the cohort-mean gain similarly rises from $+0.44\text{--}0.55\text{ pp}$ on B17 to $+1.55\text{--}1.58\text{ pp}$ on BC40 across all $6/6$ amputees with $p \le 0.00361$.

In the bottom panel of Fig.~\ref{fig:3}, the 10-subject DB5-All10 armband cohort exhibits a consistent upward shift from B17 to ABC52, where the cohort-mean accuracy gain over Gate-Only under $50\%$ corruption increases from $+2.40\text{--}2.65\text{ pp}$ on B17 to $+3.65\text{--}4.15\text{ pp}$ on ABC52. This upward shift amounts to $+1.25\text{ pp}$ under Burst with $9/10$ participants increasing, $+1.63\text{ pp}$ under Dropout with $10/10$ participants increasing, and $+1.43\text{ pp}$ under Liftoff with $9/10$ participants increasing, all with paired-$t$ $p \le 1.15 \times 10^{-4}$. Under $25\%$ corruption in Supplementary Table S5, the mean gain over Gate-Only nearly doubles from $+1.03\text{--}1.29\text{ pp}$ on B17 to $+2.05\text{--}2.32\text{ pp}$ on ABC52 with $p \le 2.59 \times 10^{-4}$.
Expanding the target repertoire from 17 gestures to 40 or 52 gestures therefore consistently amplifies the accuracy advantage of QA-CMIF over Gate-Only across both clinical amputee and intact cohorts.

\begin{figure*}[!t]
\centering
\includegraphics[width=.76\textwidth,trim=0bp 35.5bp 0bp 0bp,clip]{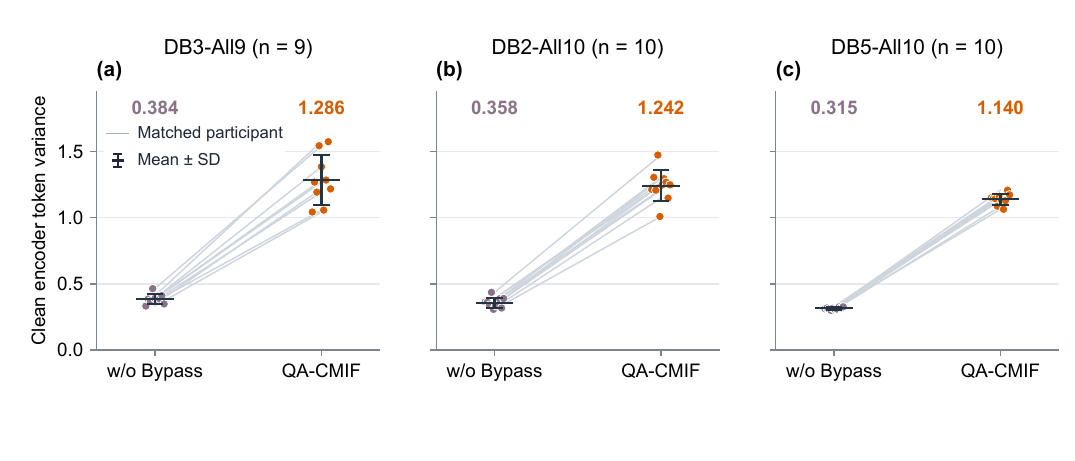}
\caption{Clean-encoder token variance $\sigma_{\text{clean}}^2$ comparing w/o Bypass and QA-CMIF on B17 across DB3-All9, DB2-All10, and DB5-All10. Markers connected by thin gray lines show three-seed means for matched participants, vertical error bars indicate the cohort mean $\pm$ SD, and values at the top of each panel report the cohort mean.}
\label{fig:4}
\end{figure*}
\begin{figure*}[!t]
\centering
\includegraphics[width=.76\textwidth,trim=0bp 44.5bp 0bp 0bp,clip]{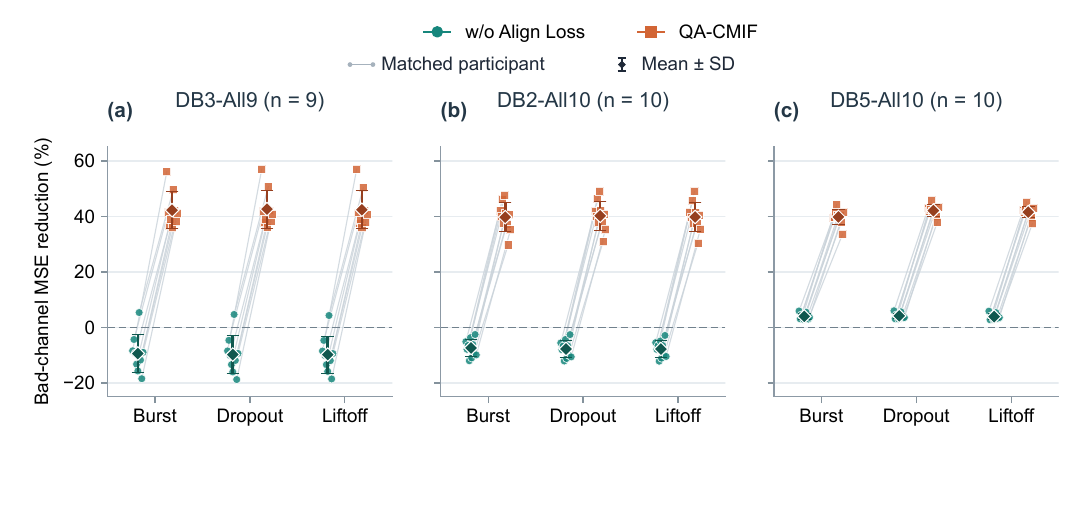}
\caption{Paired participant-level bad-channel latent MSE reduction comparing w/o Align Loss and QA-CMIF on B17 under 50\% channel corruption across DB3-All9, DB2-All10, and DB5-All10. Markers connected by thin gray lines represent three-seed mean reductions for matched participants computed via Equation~\eqref{eq:18}, and dark markers with vertical error bars indicate the cohort mean $\pm$ SD across participants.}
\label{fig:5}
\end{figure*}
\begin{figure}[!t]
\centering
\includegraphics[width=\columnwidth,trim=0bp 44bp 0bp 0bp,clip]{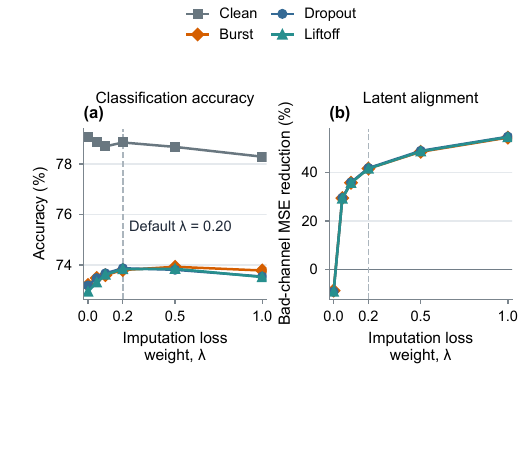}
\caption{Sensitivity of classification accuracy and bad-channel latent MSE reduction to the alignment loss weight $\lambda_{\text{impute}}$ on DB3-Core6 for B17 under clean and 50\% channel corruption conditions. Markers connected by solid lines show cohort means across the six participants, and the vertical dashed line indicates the default setting $\lambda_{\text{impute}} = 0.20$.}
\label{fig:6}
\end{figure}
\begin{figure*}[!tb]
\centering
\includegraphics[width=.80\textwidth,trim=0bp 48.5bp 0bp 0bp,clip]{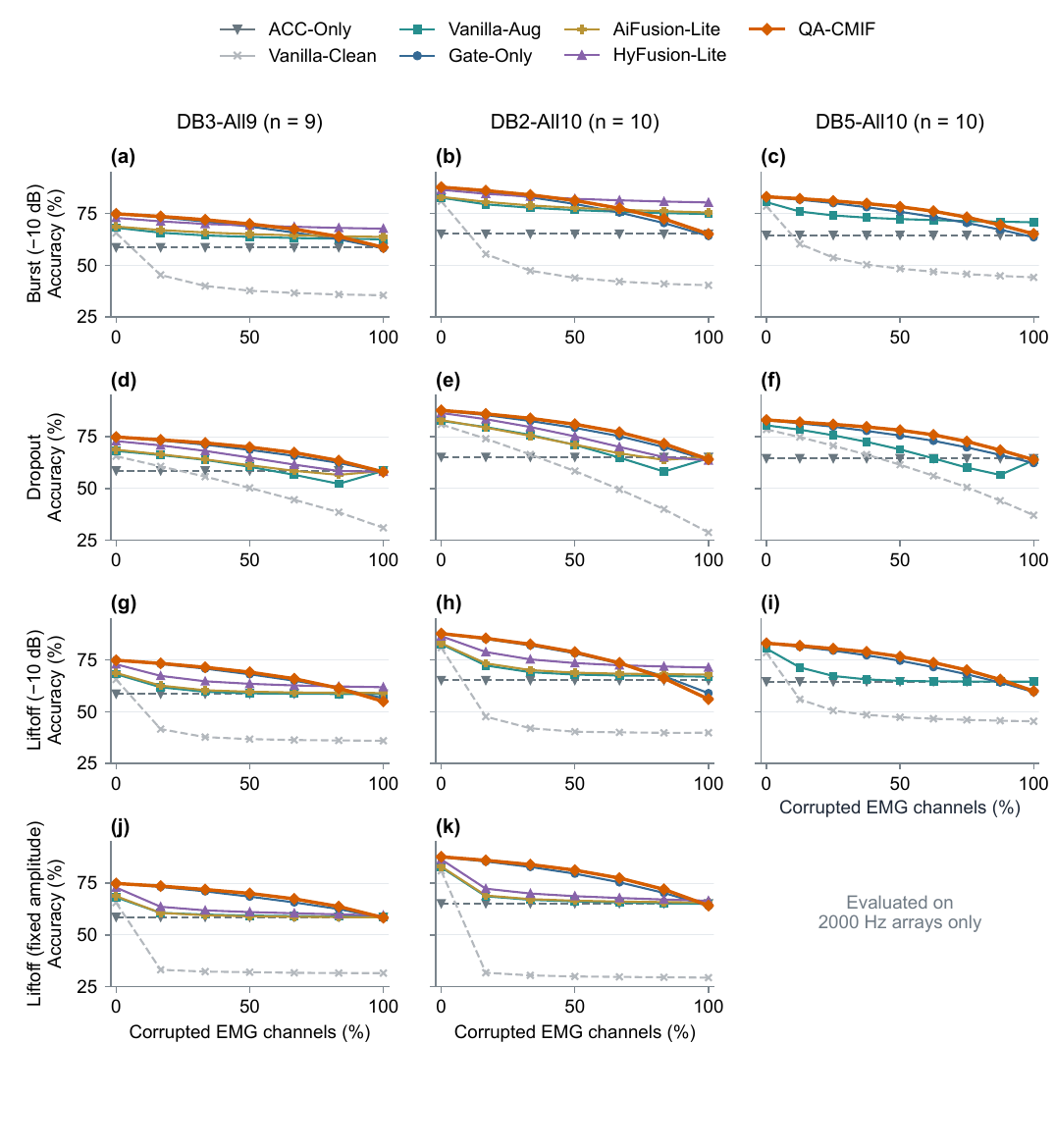}
\caption{Classification accuracy versus the percentage of corrupted sEMG channels on B17 across DB3-All9, DB2-All10, and DB5-All10. Markers along each curve represent the cohort mean across participants after averaging three seeds per participant, evaluated in steps of two corrupted channels from 0 to 12 on the 2000 Hz DB3 and DB2 arrays and from 0 to 16 on the 200 Hz DB5 armbands. Fixed-amplitude 50 Hz Liftoff was evaluated on the 2000 Hz DB3 and DB2 arrays only.}
\label{fig:7}
\end{figure*}
\begin{figure*}[!tb]
\centering
\includegraphics[width=.72\textwidth,trim=0bp 20bp 0bp 0bp,clip]{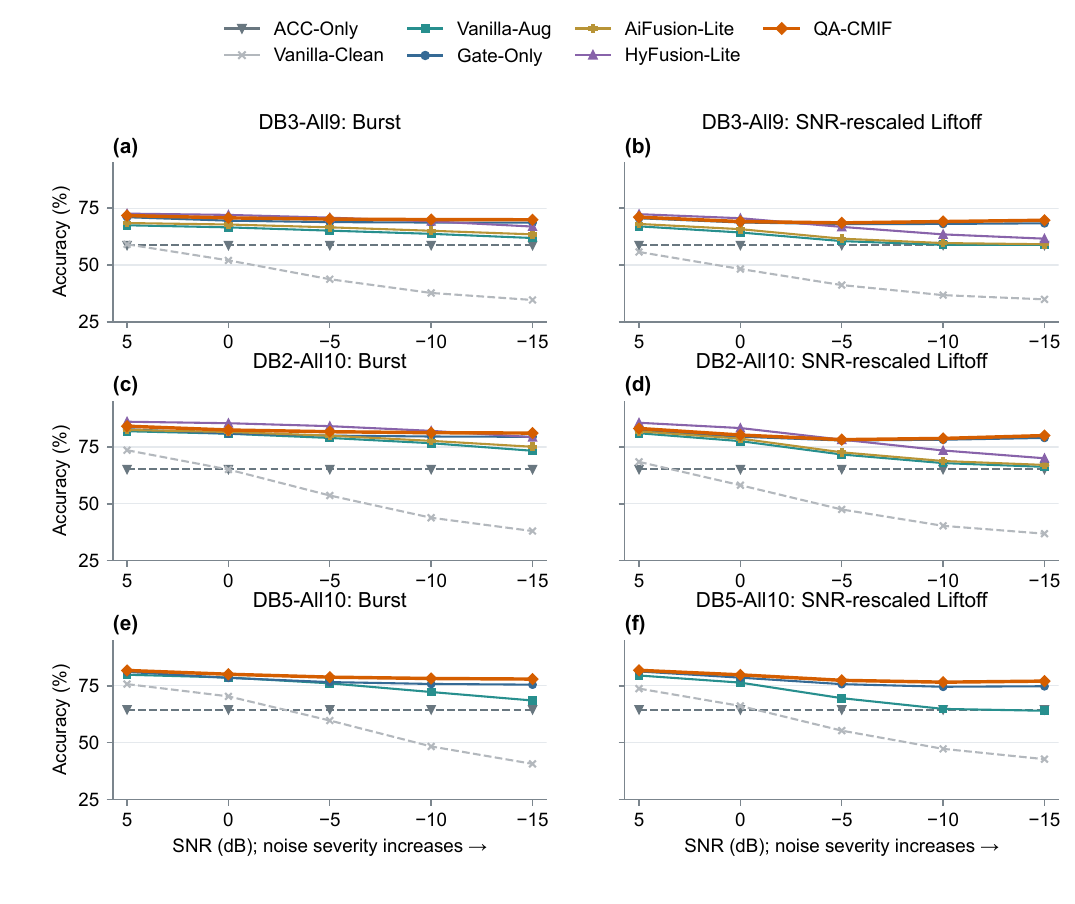}
\caption{Classification accuracy versus artifact signal-to-noise ratio on B17 under 50\% sEMG channel corruption across DB3-All9, DB2-All10, and DB5-All10. Markers along each curve represent the cohort mean across participants after averaging three seeds per participant at each evaluated SNR level.}
\label{fig:8}
\end{figure*}
\begin{table*}[!t]
\centering
\caption{Controlled Ablation Study and Internal Diagnostics on B17 (Mean $\pm$ SD)}
\label{tab:5}
\fontsize{7.5}{8.5}\selectfont
\setlength{\tabcolsep}{2.2pt}
\renewcommand{\arraystretch}{1.0}
\begin{tabular*}{\textwidth}{@{\extracolsep{\fill}}lccccccccr@{}}
\toprule
\strong{Model} & \shortstack{\strong{clean Acc}\\\strong{(\%)}} & \shortstack{\strong{Burst-$50\%$}\\\strong{(\%)}} & \shortstack{\strong{Dropout-$50\%$}\\\strong{(\%)}} & \shortstack{\strong{Liftoff-$50\%$}\\\strong{(\%)}} & \shortstack{\strong{Gate AUROC/F1}\\\strong{(Burst-$50\%$)}$^{\text{a}}$} & \shortstack{\strong{Clean}\\\strong{FAR (\%)}} & \shortstack{\strong{Clean disp.}\\\strong{$\sigma_{\text{clean}}^2$}} & \shortstack{\strong{MSE red.}\\\strong{Liftoff-$50\%$ (\%)}} & \shortstack{\strong{Learned}\\\strong{$\gamma$}} \\
\midrule
\multicolumn{10}{@{}l}{\strong{DB3-All9} ($n = 9$ amputees, $2000\text{ Hz}$)} \\
Vanilla-Aug & 68.13 $\pm$ 12.81 & 63.70 $\pm$ 14.53 & 60.49 $\pm$ 13.99 & 59.33 $\pm$ 15.16 & --- & --- & --- & --- & --- \\
Gate-Only & 74.93 $\pm$ 12.39 & 68.67 $\pm$ 13.81 & 68.65 $\pm$ 13.77 & 68.48 $\pm$ 13.82 & 0.9991 / 0.9904 & 0.41 & 1.405 $\pm$ 0.194 & 0.00 $\pm$ 0.00 & --- \\
w/o Bypass & \strong{75.23 $\pm$ 11.51} & 69.03 $\pm$ 13.34 & 68.98 $\pm$ 13.55 & 69.00 $\pm$ 13.44 & 0.9989 / 0.9863 & 0.35 & 0.384 $\pm$ 0.038 & +37.09 $\pm$ 11.28 & --- \\
w/o Align Loss & 74.93 $\pm$ 12.31 & 69.46 $\pm$ 13.41 & 69.50 $\pm$ 13.46 & 69.39 $\pm$ 13.48 & 0.9992 / 0.9908 & 0.41 & 1.364 $\pm$ 0.185 & $-$9.79 $\pm$ 6.70 & 0.655 $\pm$ 0.050 \\
\strong{QA-CMIF} & 74.85 $\pm$ 12.63 & \strong{69.97 $\pm$ 13.48} & \strong{69.97 $\pm$ 13.48} & \strong{70.04 $\pm$ 13.49} & 0.9992 / 0.9898 & 0.43 & 1.286 $\pm$ 0.189 & +42.43 $\pm$ 6.90 & 0.625 $\pm$ 0.037 \\
\midrule
\multicolumn{10}{@{}l}{\strong{DB2-All10} ($n = 10$ intact, $2000\text{ Hz}$)} \\
Vanilla-Aug & 82.63 $\pm$ 6.55 & 76.67 $\pm$ 8.11 & 71.03 $\pm$ 7.05 & 66.15 $\pm$ 6.87 & --- & --- & --- & --- & --- \\
Gate-Only & 87.58 $\pm$ 3.75 & 79.61 $\pm$ 5.20 & 79.28 $\pm$ 5.31 & 79.57 $\pm$ 5.33 & 0.9991 / 0.9897 & 0.41 & 1.350 $\pm$ 0.131 & 0.00 $\pm$ 0.00 & --- \\
w/o Bypass & \strong{88.84 $\pm$ 3.47} & 80.61 $\pm$ 5.06 & 80.27 $\pm$ 4.84 & 80.37 $\pm$ 4.94 & 0.9979 / 0.9733 & 0.44 & 0.358 $\pm$ 0.039 & +36.89 $\pm$ 10.30 & --- \\
w/o Align Loss & 87.74 $\pm$ 3.50 & 80.75 $\pm$ 4.97 & 80.71 $\pm$ 5.09 & 80.91 $\pm$ 5.03 & 0.9992 / 0.9904 & 0.42 & 1.306 $\pm$ 0.128 & $-$7.78 $\pm$ 3.03 & 0.628 $\pm$ 0.024 \\
\strong{QA-CMIF} & 87.72 $\pm$ 3.66 & \strong{81.28 $\pm$ 4.72} & \strong{81.03 $\pm$ 4.78} & \strong{81.26 $\pm$ 4.86} & 0.9989 / 0.9868 & 0.42 & 1.242 $\pm$ 0.118 & +39.84 $\pm$ 5.20 & 0.595 $\pm$ 0.032 \\
\midrule
\multicolumn{10}{@{}l}{\strong{DB5-All10} ($n = 10$ intact, $200\text{ Hz}$)} \\
Vanilla-Aug & 80.49 $\pm$ 4.30 & 72.26 $\pm$ 5.74 & 68.71 $\pm$ 4.67 & 63.63 $\pm$ 6.96 & --- & --- & --- & --- & --- \\
Gate-Only & 83.04 $\pm$ 4.18 & 75.78 $\pm$ 4.97 & 75.50 $\pm$ 4.86 & 75.46 $\pm$ 4.83 & 0.9959 / 0.9725 & 0.42 & 1.237 $\pm$ 0.034 & 0.00 $\pm$ 0.00 & --- \\
w/o Bypass & \strong{84.11 $\pm$ 4.58} & 76.59 $\pm$ 4.76 & 76.77 $\pm$ 4.71 & 76.84 $\pm$ 4.60 & 0.9910 / 0.9429 & 0.51 & 0.315 $\pm$ 0.008 & +43.37 $\pm$ 4.94 & --- \\
w/o Align Loss & 83.14 $\pm$ 4.12 & 77.08 $\pm$ 4.59 & 76.98 $\pm$ 4.46 & 77.03 $\pm$ 4.46 & 0.9960 / 0.9737 & 0.43 & 1.200 $\pm$ 0.032 & +3.91 $\pm$ 1.02 & 0.448 $\pm$ 0.019 \\
\strong{QA-CMIF} & 83.08 $\pm$ 4.02 & \strong{78.18 $\pm$ 4.42} & \strong{78.01 $\pm$ 4.34} & \strong{78.11 $\pm$ 4.37} & 0.9953 / 0.9694 & 0.43 & 1.140 $\pm$ 0.041 & +41.59 $\pm$ 2.05 & 0.451 $\pm$ 0.018 \\
\bottomrule
\end{tabular*}
\vspace{2pt}
\begin{flushleft}
\scriptsize $^{\text{a}}$Under Dropout-$50\%$ and Liftoff-$50\%$, QA-CMIF achieves Gate AUROC $\ge 0.9995$ and Gate F1 $\ge 0.9960$ across all three cohorts. Figs.~\ref{fig:4} and \ref{fig:5} illustrate the paired clean representation variance $\sigma_{\text{clean}}^2$ and bad-channel MSE reduction $\text{MSE}_{\text{red}}(\%)$ for the corresponding ablation variants. Accuracy values are in $\%$ (bold indicates highest per cohort and test condition); diagnostic columns are descriptive and unbolded. Detailed w/o Stop-Grad results are provided in Supplementary Table S4.
\end{flushleft}
\end{table*}
\subsection{Controlled Ablation Study and Latent Diagnostics}

To verify the individual contributions of channel-independent soft gating, the confidence-weighted residual bypass, the bad-channel alignment loss, and gate stop-gradient detachment,
we conducted a controlled ablation study on B17 across DB3-All9, DB2-All10, and DB5-All10 in Table V, Supplementary Table S4, and Figs.~\ref{fig:4} and \ref{fig:5}. Alongside classification accuracy and gate detection metrics, Fig.~\ref{fig:4} compares clean-encoder token dispersion $\sigma_{\text{clean}}^2$, defined as the mean per-dimension variance of uncorrupted sEMG tokens across test windows, between w/o Bypass and QA-CMIF, while Fig.~\ref{fig:5} compares within-model bad-channel latent MSE reduction between w/o Align Loss and QA-CMIF over $\mathcal{B} = \{(b,c) \mid m_{bc}=0\}$ relative to the same model's attenuated tokens given by

\begin{equation}
\begin{aligned}
&\text{MSE}_{\text{red}}(\%) =\\[-2pt]
&\left(1-\frac{\sum_{(b,c)\in\mathcal{B}}\|H_{\text{emg},bc}-z_{\text{clean},bc}\|_2^2}
{\sum_{(b,c)\in\mathcal{B}}\|\alpha_{bc}z_{\text{emg},bc}-z_{\text{clean},bc}\|_2^2}\right)\times100\%
\end{aligned}
\label{eq:18}
\end{equation}

In Table V, replacing the early cross-channel mixing of Vanilla-Aug with channel-independent encoding and multi-statistic soft gating in Gate-Only increases clean accuracy by $+2.55\text{ to }{+}6.80\text{ pp}$ and improves $50\%$-corruption accuracy by $+4.97\text{ to }{+}9.15\text{ pp}$ on DB3-All9 with $9/0$ subject wins and $p_{\text{raw}} \le 3.52 \times 10^{-5}$, $+2.94\text{ to }{+}13.43\text{ pp}$ on DB2-All10, and $+3.53\text{ to }{+}11.83\text{ pp}$ on DB5-All10. Across all three cohorts, ranking intact channels with $m_c = 1$ against corrupted channels with $m_c = 0$ via $\alpha_c$ using the descriptors in Equations (2) through (5) yields a gate AUROC of $0.9953\text{--}1.0000$, a fault-detection F1 score of $0.9694\text{--}0.9980$ at the $\alpha_c < 0.5$ threshold \cite{ref9}, \cite{ref12}, and a clean false alarm rate, denoted FAR and defined as the fraction of intact channels flagged on uncorrupted windows, of $0.35\%\text{--}0.51\%$ across all gated variants and at most $0.43\%$ for QA-CMIF.

Comparing QA-CMIF against the w/o Bypass ablation, which sets $H_{\text{emg}} = \tilde{Z}_{\text{imp}}$, shows that clean-encoder token variance $\sigma_{\text{clean}}^2$ in Table V and Fig.~\ref{fig:4} drops by more than $70\%$ when the residual bypass is removed, falling from $1.286$ to $0.384$ on DB3-All9, from $1.242$ to $0.358$ on DB2-All10, and from $1.140$ to $0.315$ on DB5-All10. Although w/o Bypass achieves $+0.38\text{ to }{+}1.12\text{ pp}$ higher accuracy on clean windows, QA-CMIF outperforms w/o Bypass across all three $50\%$-corruption modes by $+0.93\text{ to }{+}1.05\text{ pp}$ on DB3-All9 with $9/9$ subject wins and $p_{\text{raw}} \le 3.33 \times 10^{-4}$, $+0.68\text{ to }{+}0.90\text{ pp}$ on DB2-All10 with $p_{\text{raw}} \le 0.0421$, and $+1.24\text{ to }{+}1.59\text{ pp}$ on DB5-All10 with $10/10$ wins and $p_{\text{raw}} \le 9.80 \times 10^{-5}$ under unadjusted paired-$t$ tests.

As shown by the paired participant comparisons in Fig.~\ref{fig:5} and Table V, removing the bad-channel alignment loss in w/o Align Loss with $\lambda_{\text{impute}} = 0$ leaves the learned scale $\gamma$ at $0.628\text{--}0.655$ on DB2 and DB3 and $0.448$ on DB5, yet bad-channel MSE reduction remains negative at $-9.41\%\text{ to }{-}9.79\%$ on DB3 and $-7.38\%\text{ to }{-}7.78\%$ on DB2, and reaches only $+3.91\%\text{ to }{+}4.17\%$ on DB5. Enabling $\mathcal{L}_{\text{impute}}$ in QA-CMIF shifts every participant upward in Fig.~\ref{fig:5}, achieving positive bad-channel MSE reductions of $39.75\%\text{--}42.65\%$ across all B17 cohorts and $48.53\%\text{--}50.80\%$ on DB5 ABC52 in Supplementary Table S4, while improving $50\%$-corruption accuracy over w/o Align Loss by $+0.47\text{ to }{+}0.66\text{ pp}$ on DB3 with $p_{\text{raw}} \le 0.0184$ on Burst and Liftoff, $+0.32\text{ to }{+}0.53\text{ pp}$ on DB2 with $p_{\text{raw}} \le 0.0423$ on Burst and Liftoff, and $+1.03\text{ to }{+}1.10\text{ pp}$ on DB5 with $10/10$ wins and $p_{\text{raw}} \le 4.34 \times 10^{-5}$. Finally, to evaluate whether detaching $\bar{\boldsymbol{\alpha}}$ inside the imputer protects gate calibration, Supplementary Table S4 compares QA-CMIF with w/o Stop-Grad; removing stop-gradient detachment slightly increases Burst-50\% accuracy by $+0.40\text{ pp}$ on DB3 with $p_{\text{raw}} = 0.0070$, $+0.54\text{ pp}$ on DB2 with $p_{\text{raw}} = 7.0 \times 10^{-4}$ and full-table $p_{\text{Holm}} = 0.0322$, and $+0.10\text{ pp}$ on DB5, with Dropout-50\% and Liftoff-50\% changes from QA-CMIF within $[-0.14, +0.10]\text{ pp}$, but degrades gate calibration on uncorrupted windows by raising clean FAR on $2000\text{ Hz}$ arrays from $0.42\%\text{--}0.43\%$ in QA-CMIF to $0.58\%\text{--}0.72\%$.
Taken together, these ablation and diagnostic results confirm that each architectural and objective component fulfills its designated role, with channel-independent gating suppressing corrupted inputs, the residual bypass preserving clean token dispersion, the alignment loss driving positive latent MSE reduction on corrupted channels, and stop-gradient detachment maintaining low clean false alarm rates.

\subsection{Degradation Sweeps Over Corrupted Channel Count and SNR}

Beyond fixed $25\%$ and $50\%$ corruption levels, practical deployment requires characterizing model behavior across the full continuum of spatial fault densities and artifact intensities.
Fig.~\ref{fig:7} sweeps the corrupted sEMG channel count $K \in \{0, 2, \dots, 12\}$ across all seven models on DB3-All9 and DB2-All10 and $K \in \{0, 2, \dots, 16\}$ across the five core models on DB5-All10, while Fig.~\ref{fig:8} sweeps the artifact signal-to-noise ratio $\text{SNR}_{\text{dB}} \in \{+5, 0, -5, -10, -15\}\text{ dB}$ at $50\%$ channel corruption with $K = C_e/2$.
In the $K$-sweeps of Fig.~\ref{fig:7}, corrupting even $K = 2$ channels under fixed-amplitude Liftoff reduces Vanilla-Clean accuracy from $65.58\%$ to $33.16\%$ on DB3 and from $80.92\%$ to $31.75\%$ on DB2, and to $41.60\%$ and $47.62\%$ under $-10\text{ dB}$ SNR-rescaled Liftoff. Across $K \in [2, 8]$ on the 12-channel arrays and $K \in [2, 12]$ on the 16-channel armbands, QA-CMIF remains consistently above Gate-Only, Vanilla-Aug, and AiFusion-Lite, with Burst gains over Gate-Only peaking at $K = 8/12$ at $+1.66\text{ pp}$ on DB3 and $+2.04\text{ pp}$ on DB2, and DB5 Dropout gains peaking at $K = 10/16$ at $+2.88\text{ pp}$ and $K = 12/16$ at $+2.85\text{ pp}$. Compared with HyFusion-Lite, QA-CMIF maintains higher accuracy across $K \in [2, 8]$ under Dropout and Liftoff, whereas HyFusion-Lite attains higher accuracy under Burst on DB2 across $K \ge 2$ and on DB3 at $K \ge 8$. As $K$ approaches $C_e$, both QA-CMIF and Gate-Only converge toward the ACC-Only baseline, and at complete 12-channel SNR-rescaled Liftoff with $K = 12$, QA-CMIF trails Gate-Only by $1.54\text{ pp}$ on DB3 and $2.92\text{ pp}$ on DB2.

In the $\text{SNR}_{\text{dB}}$ sweeps of Fig.~\ref{fig:8} at $50\%$ channel corruption with $K = C_e/2$, the ungated models Vanilla-Clean, Vanilla-Aug, AiFusion-Lite, and HyFusion-Lite decline monotonically as noise intensifies from $+5\text{ dB}$ to $-15\text{ dB}$, with rescaled Liftoff on DB2 reducing HyFusion-Lite accuracy from $85.66\%$ at $+5\text{ dB}$ to $70.05\%$ at $-15\text{ dB}$. In contrast, QA-CMIF and Gate-Only maintain a flat accuracy plateau across $\text{SNR}_{\text{dB}} \in [-5, -15]\text{ dB}$, where Burst accuracy varies by at most $0.28\text{ pp}$ on DB3, $0.60\text{ pp}$ on DB2, and $1.13\text{ pp}$ on DB5. From $-5\text{ dB}$ to $-15\text{ dB}$, rescaled Liftoff accuracy changes by $+1.15\text{ pp}$ and $+0.38\text{ pp}$ on DB3, $+1.79\text{ pp}$ and $+1.14\text{ pp}$ on DB2, and $-0.34\text{ pp}$ and $-0.92\text{ pp}$ on DB5 for QA-CMIF and Gate-Only, respectively. Across $-5$, $-10$, and $-15\text{ dB}$, QA-CMIF outperforms Gate-Only by $+1.29\text{ to }{+}2.45\text{ pp}$ under Burst and $+0.37\text{ to }{+}2.26\text{ pp}$ under rescaled Liftoff across all three cohorts. At $+5\text{ dB}$ on DB3 and DB2, HyFusion-Lite starts at a higher accuracy than QA-CMIF, whereas HyFusion-Lite falls below QA-CMIF under Liftoff at $-5\text{ dB}$ and below on DB3 and at $-10\text{ dB}$ and below on DB2. Under Burst, HyFusion-Lite falls below QA-CMIF at $-10\text{ dB}$ and below on DB3 and at $-15\text{ dB}$ on DB2.
Overall, the degradation sweeps demonstrate that QA-CMIF sustains a consistent accuracy lead over Gate-Only and ungated baselines across moderate-to-severe channel corruption fractions $K \in [2, 8]$ on 12-channel arrays and $K \in [2, 12]$ on 16-channel armbands, while exhibiting the limited accuracy variations reported above across $\text{SNR}_{\text{dB}} \in [-5, -15]\text{ dB}$.

\subsection{Computational Efficiency, Inference Latency, and Hyperparameter Sensitivity}

Practical deployment on wearable prosthetic controllers requires a lightweight memory footprint, low single-window latency, and insensitivity to hyperparameter tuning.
Table VI benchmarks parameter counts, multiply--accumulate operations, denoted MACs, and batch-one forward inference latency over 1,000 timed runs after 100 warm-up iterations on an NVIDIA GeForce RTX 5090 GPU and a single-thread x86\_64 CPU core, while Fig.~\ref{fig:6} sweeps the bad-channel alignment weight $\lambda_{\text{impute}} \in \{0.00, 0.05, 0.10, 0.20, 0.50, 1.00\}$ on DB3-Core6 B17.
Under the $2000\text{ Hz}$ setup with 12-EMG and 3-ACC inputs and $T = 400$ in Table VI, QA-CMIF requires 174,867 parameters and $2.63\text{M}$ MACs, using $+0.18\%$ more parameters but $35.4\%$ fewer MACs than Vanilla-Clean and Vanilla-Aug, while adding 35,073 parameters and $0.42\text{M}$ MACs over Gate-Only. Single-thread CPU mean latency at $2000\text{ Hz}$ is $0.545 \pm 0.012\text{ ms}$ for Vanilla-Aug, $0.987 \pm 0.047\text{ ms}$ for Gate-Only, and $1.358 \pm 0.111\text{ ms}$ for QA-CMIF, with a 99th-percentile latency, denoted P99, of $1.443\text{ ms}$ on CPU and $1.615\text{ ms}$ on GPU, occupying less than $2.9\%$ of the $50\text{ ms}$ window stride \cite{ref1}. Relative to HyFusion-Lite and AiFusion-Lite at $2000\text{ Hz}$, QA-CMIF reduces parameters by $55.2\%$ and $53.5\%$, cuts MACs by $80.9\%$ and $73.4\%$, and lowers CPU mean latency by $49.3\%$ and $44.8\%$, respectively; under the $200\text{ Hz}$ DB5 setup, QA-CMIF requires 184,915 parameters, $2.45\text{M}$ MACs, and $1.338\text{ ms}$ CPU latency, saving $46.4\%\text{--}50.1\%$ in parameters and $71.9\%\text{--}75.6\%$ in MACs.

In the sensitivity scan across $\lambda_{\text{impute}} \in \{0.00, 0.05, 0.10, 0.20, 0.50, 1.00\}$ on DB3-Core6 B17 with $n = 6$ in Fig.~\ref{fig:6}, setting $\lambda_{\text{impute}} = 0.00$ yields negative bad-channel MSE reductions of $-8.67\%$ to $-9.12\%$ and $50\%$-corruption accuracies of $72.95\%\text{--}73.21\%$. Increasing $\lambda_{\text{impute}}$ to $0.05$ raises bad-channel MSE reduction to $29.24\%\text{--}29.56\%$. Across $\lambda_{\text{impute}} \in [0.10, 0.50]$, bad-channel MSE reduction increases monotonically from $35.71\%$ to $48.94\%$ while $50\%$-corruption accuracy remains within $73.60\%\text{--}73.93\%$, varying by at most $0.25\text{ pp}$ around the default $\lambda_{\text{impute}} = 0.20$ across Burst, Dropout, and Liftoff. At $\lambda_{\text{impute}} = 1.00$, bad-channel MSE reduction reaches $54.39\%\text{--}54.75\%$ while clean accuracy decreases from $78.85\%$ at $\lambda_{\text{impute}} = 0.20$ to $78.29\%$.
QA-CMIF thus operates with fewer than $29\%$ of the MACs of adapted multimodal baselines, completes single-window CPU inference within $2.9\%$ of the $50\text{ ms}$ stride budget, and maintains stable accuracy across a fivefold range of $\lambda_{\text{impute}} \in [0.10, 0.50]$.

\begin{table*}[!t]
\centering\begingroup
\fontsize{7.5}{8.5}\selectfont
\setlength{\tabcolsep}{3.5pt}
\renewcommand{\arraystretch}{0.95}
\caption{COMPUTATIONAL COMPLEXITY AND SINGLE-WINDOW FORWARD INFERENCE LATENCY ACROSS $2000\text{ Hz}$ AND $200\text{ Hz}$ ACQUISITION REGIMES (17 CLASSES, $\text{BATCH} = 1$)}
\label{tab:VII}
\begin{tabular}{@{}lccccccc@{}}
\toprule
Model & Params & Rel. params & MACs & FLOPs & Rel. MACs & \shortstack{CPU mean $\pm$ SD\\(P99, ms)} & \shortstack{GPU mean $\pm$ SD\\(P99, ms)}\\\midrule
\midrule\multicolumn{8}{@{}l}{\strong{$2000\text{ Hz}$ Array}  (DB3 / DB2,   $12\text{-EMG} + 3\text{-ACC}$,   $T = 400$)}\\
ACC-Only & 81,233 & $0.46\times$ & 1.46M & 2.92M & $0.55\times$ & 0.260 $\pm$ 0.010 (0.299) & 0.271 $\pm$ 0.061 (0.342)\\
Vanilla-Clean & 174,545 & $1.00\times$ & 4.07M & 8.14M & $1.55\times$ & 0.553 $\pm$ 0.012 (0.602) & 0.410 $\pm$ 0.004 (0.422)\\
Vanilla-Aug & 174,545 & $1.00\times$ & 4.07M & 8.14M & $1.55\times$ & 0.545 $\pm$ 0.012 (0.595) & 0.410 $\pm$ 0.003 (0.422)\\
Gate-Only (Control) & 139,794 & $0.80\times$ & 2.21M & 4.42M & $0.84\times$ & 0.987 $\pm$ 0.047 (1.056) & 0.800 $\pm$ 0.022 (0.826)\\
\strong{QA-CMIF} & \strong{174,867} & \strong{$1.00\times$} & \strong{2.63M} & \strong{5.26M} & \strong{$1.00\times$} & \strong{1.358 $\pm$ 0.111 (1.443)} & \strong{1.446 $\pm$ 0.386 (1.615)}\\
AiFusion-Lite \cite{ref13} & 376,115 & $2.15\times$ & 9.90M & 19.79M & $3.76\times$ & 2.458 $\pm$ 0.255 (2.563) & 3.800 $\pm$ 0.261 (3.908)\\
HyFusion-Lite \cite{ref20} & 390,486 & $2.23\times$ & 13.75M & 27.51M & $5.23\times$ & 2.678 $\pm$ 0.498 (2.812) & 3.710 $\pm$ 0.328 (3.792)\\
\midrule\multicolumn{8}{@{}l}{\strong{$200\text{ Hz}$ Armband}  (DB5,   $16\text{-EMG} + 3\text{-ACC}$,   $T = 40$)}\\
Vanilla-Clean & 156,625 & $0.85\times$ & 2.39M & 4.78M & $0.98\times$ & 0.464 $\pm$ 0.149 (1.283) & 0.403 $\pm$ 0.150 (0.625)\\
Vanilla-Aug & 156,625 & $0.85\times$ & 2.39M & 4.78M & $0.98\times$ & 0.448 $\pm$ 0.104 (0.524) & 0.403 $\pm$ 0.149 (0.894)\\
Gate-Only (Control) & 149,394 & $0.81\times$ & 1.88M & 3.76M & $0.77\times$ & 0.964 $\pm$ 0.250 (1.869) & 0.813 $\pm$ 0.224 (1.398)\\
\strong{QA-CMIF} & \strong{184,915} & \strong{$1.00\times$} & \strong{2.45M} & \strong{4.89M} & \strong{$1.00\times$} & \strong{1.338 $\pm$ 0.303 (2.312)} & \strong{1.424 $\pm$ 0.442 (2.345)}\\
AiFusion-Lite \cite{ref13} & 370,611 & $2.00\times$ & 8.71M & 17.41M & $3.56\times$ & 2.433 $\pm$ 0.389 (4.036) & 3.846 $\pm$ 0.198 (4.116)\\
HyFusion-Lite \cite{ref20} & 344,918 & $1.87\times$ & 10.01M & 20.02M & $4.09\times$ & 2.523 $\pm$ 0.627 (5.697) & 3.684 $\pm$ 0.673 (7.198)\\
\bottomrule\end{tabular}
\par\vspace{2pt}
\begin{minipage}{\textwidth}\scriptsize
$^{\text{a}}$ Benchmarked in PyTorch 2.8.0+cu128 (float32, single-thread x86\_64 host CPU core and NVIDIA GeForce RTX 5090 GPU) over 1,000 forward passes after 100 warm-up iterations; MACs count Conv1d, Linear, and attention projections/matrix multiplications ($\text{FLOPs} = 2\times\text{MACs}$ convention, excluding activations, pooling, and normalization); Rel. denotes the ratio to QA-CMIF, and P99 denotes the 99th percentile. Latency values are reported in milliseconds (ms).
\end{minipage}\endgroup
\end{table*}
\FloatBarrier
\section{Discussion}

\subsection{Interpreting the Gains and Trade-Offs Across Corruption Types and Baselines}

The results across Ninapro DB3, DB2, and DB5 demonstrate that reliable myoelectric--inertial decoding under unknown partial channel corruption benefits from coupling early fault isolation with cross-modal latent repair. First, replacing early cross-channel mixing convolutions with channel-independent encoding and multi-statistic soft gating prevents localized high-amplitude artifacts from contaminating intact electrode representations at the input stage. Second, whereas gate-only attenuation suppresses degraded channels toward zero and discards inter-muscular co-activation cues \cite{ref31}, QA-CMIF queries the intact reference IMU alongside surviving high-confidence sEMG tokens while preserving uncorrupted representations through a confidence-weighted residual bypass. This complementary design is particularly effective on fine-grained 40- and 52-class finger and grasp taxonomies \cite{ref6}, where subtle spatial synergy across forearm musculature separates closely related movements, and achieves fault-tolerant decoding in a compact latent space without the computational overhead of raw waveform reconstruction \cite{ref11}, \cite{ref22}.

Comparing QA-CMIF with HyFusion-Lite \cite{ref20} also clarifies the complementary behavior of explicit quality gating and high-capacity spatial filtering. Under mild $+5\text{ dB}$ perturbations or band-limited burst noise on intact subjects in DB2, the $5.23\times$ heavier multi-branch convolutions of HyFusion-Lite can partially smooth low-amplitude zero-mean oscillations across redundant intact musculature, whereas binary gate supervision encourages QA-CMIF to attenuate perturbed channels even when residual myoelectric cues remain. Under open-circuit dropouts, non-zero-mean lift-off shocks, severe noise from $-5\text{ dB}$ to $-15\text{ dB}$, or sparser transradial amputee recordings, however, dense spatial filters propagate structural distortions across all feature maps, whereas explicit gating and cross-modal latent repair maintain a stable accuracy plateau at a fraction of the computational cost.

\begin{figure}[!htb]
\centering
\includegraphics[width=\columnwidth,trim=0bp 52bp 0bp 0bp,clip]{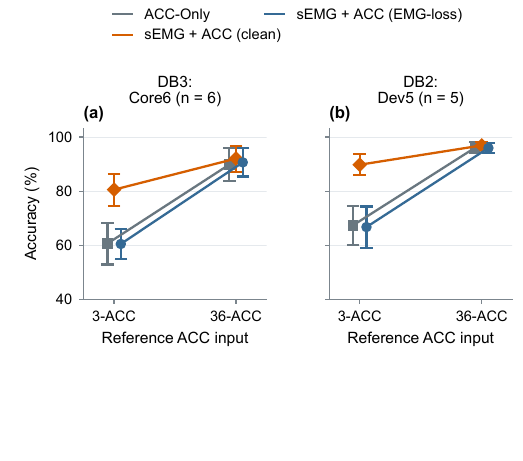}
\caption{Diagnostic comparison of single-reference 3-ACC and multi-sensor 36-ACC inertial inputs on DB3-Core6 and DB2-Dev5 for B17 under clean, EMG-loss, and inertial-only conditions. Markers and vertical error bars indicate the cohort mean $\pm$ SD across participants.}
\label{fig:9}
\end{figure}
\subsection{Comparison of Single-Reference and Multi-Sensor Inertial Configurations}

In our primary DB3 and DB2 experiments, we paired the 12-channel sEMG array with a single reference tri-axial accelerometer in the 3-ACC configuration rather than using all 12 co-located accelerometers in the 36-ACC configuration packaged in the Delsys Trigno sensors. Beyond matching practical prosthetic sockets and consumer armbands that house a single central IMU \cite{ref3}, \cite{ref14}, this protocol choice is supported by an initial diagnostic comparison on DB3-Core6 and DB2-Dev5 on B17 in Fig.~\ref{fig:9}.

As shown in Fig.~\ref{fig:9}, feeding all 36 co-located acceleration channels across the forearm and upper arm \cite{ref6} allows the purely inertial ACC-Only baseline to reach $89.86\%$ on DB3 and $96.06\%$ on DB2 without any sEMG input, leaving less than a $1.5\text{ pp}$ margin between clean and complete EMG-loss conditions because distributed multi-point accelerometers capture local surface-tilt and mechanical deformation cues \cite{ref8} that nearly saturate 17-class classification. Under the realistic single-reference 3-ACC setup, ACC-Only drops to $60.64\%$ on DB3 and $67.40\%$ on DB2, leaving an approximately $20\text{--}22\text{ pp}$ myoelectric performance margin that makes fault-tolerant sEMG processing essential.

\begin{figure*}[!tbp]
\centering
\includegraphics[width=.76\textwidth,trim=0bp 45bp 0bp 0bp,clip]{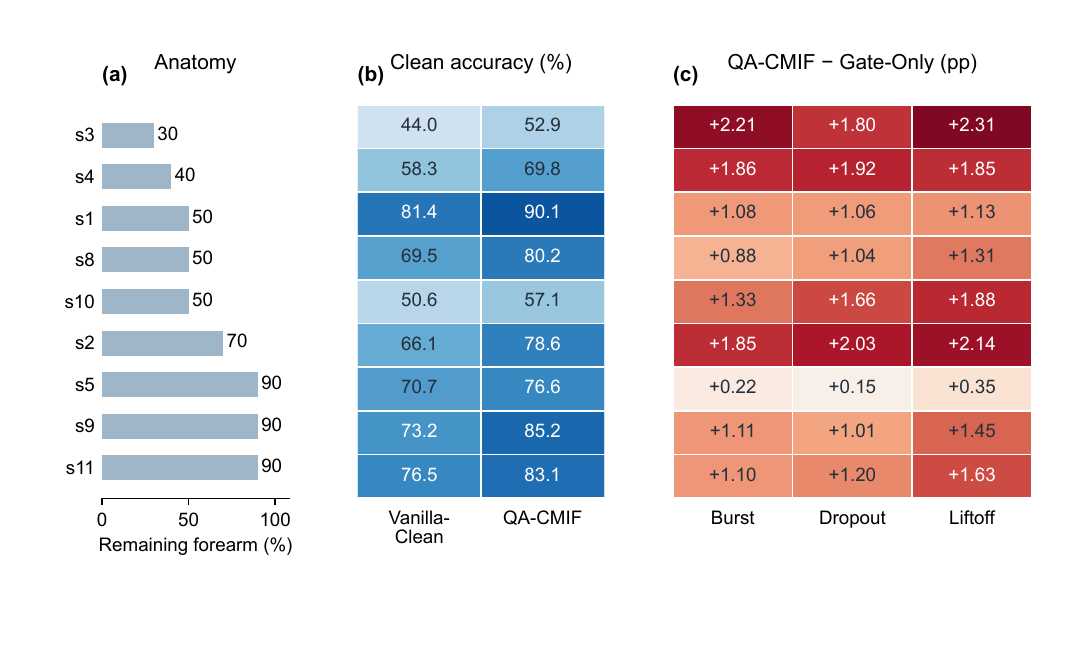}
\caption{Clinical anatomical profile and participant-level performance across the nine transradial amputees in DB3-All9 on B17, ordered by remaining forearm percentage. Panel (a) shows remaining forearm percentage, panel (b) reports three-seed mean clean accuracy for Vanilla-Clean and QA-CMIF, and panel (c) reports the three-seed mean accuracy gain of QA-CMIF over Gate-Only under 50\% channel corruption.}
\label{fig:10}
\end{figure*}
\subsection{Exploratory Analysis of Residual Forearm Length in Transradial Amputees}

Clinical transradial amputation introduces substantial anatomical heterogeneity due to variations in amputation level, residual muscle atrophy, and scar tissue \cite{ref6}, \cite{ref7}. Across the nine amputees in DB3-All9, remaining forearm length spans $30\%\text{--}90\%$ and clean baseline accuracy ranges from $52.90\%$ to $90.10\%$, as shown in Supplementary Table S1 and Fig.~\ref{fig:10}.

Despite this anatomical dispersion, QA-CMIF delivers positive accuracy gains over both Gate-Only and Vanilla-Aug across all $9/9$ amputees under every $50\%$ corruption regime in Fig.~\ref{fig:10}. Notably, even subject s3, who retains only $30\%$ of the residual forearm, achieves consistent net gains of $+1.80\text{ to }{+}2.31\text{ pp}$ over Gate-Only across the three corruption modes, indicating that cross-modal latent imputation remains effective when residual musculature is severely limited, although larger clinical cohorts will be needed for definitive subgroup stratification.

\subsection{Operating Boundaries, Limitations, and Future Work}

Several operating boundaries of this study point to directions for future work. First, because the cross-attention imputer reconstructs degraded electrode representations from the reference IMU token and surviving sEMG tokens, its restorative capacity depends on available context; when nearly all sEMG channels are simultaneously corrupted at $K \ge 10/12$ in Fig.~\ref{fig:7} or when the reference inertial stream is lost under ACC-loss in Tables III and IV, QA-CMIF naturally converges toward the gate-only and single-modality baselines.

Second, while training the quality gate with synthetic Burst, Liftoff, and Dropout masks enables reproducible benchmarking without manual annotations, future work will incorporate temporal state tracking and self-supervised test-time gate adaptation to accommodate gradual cross-day impedance drift and electrode shift during long-term wear \cite{ref5}, \cite{ref8}. Third, because all experiments were conducted offline under a within-subject cross-repetition split, deploying the lightweight QA-CMIF architecture on embedded microcontroller hardware for online, closed-loop prosthetic control with amputee users remains an important next step \cite{ref1}, \cite{ref9}.

\FloatBarrier
\section{Conclusion}

This paper presented QA-CMIF, a quality-aware cross-modal latent imputation and fusion framework for sEMG–inertial hand gesture recognition under unknown partial channel corruption. By coupling early channel-independent encoding and multi-statistic continuous soft gating with a confidence-weighted healthy residual bypass and a cross-attention imputer supervised by a bad-channel alignment loss, QA-CMIF blocks early cross-channel artifact contagion while recovering spatial myoelectric synergy in latent space. Within-subject evaluations on 29 participants across Ninapro DB3, DB2, and DB5 spanning 17--52 gesture classes demonstrated that under 50\% channel corruption, QA-CMIF improves accuracy by $1.29\text{--}4.15\text{ pp}$ over gate-only attenuation and $4.61\text{--}19.85\text{ pp}$ over corruption-augmented conventional fusion while requiring only $0.175\text{--}0.185\text{M}$ parameters, $2.45\text{--}2.63\text{M}$ MACs, and $1.44\text{ ms}$ single-thread CPU P99 latency. These findings suggest that repairing degraded channels in latent space, rather than reconstructing raw waveforms, is a practical route toward resilient wearable myoelectric interfaces. Future work will extend this framework to unsupervised cross-day drift adaptation and online closed-loop prosthetic control. The Ninapro DB2, DB3, and DB5 benchmarks analyzed in this study are publicly available from the official Ninapro repository, and our source code, deterministic corruption generators, and evaluation scripts will be released publicly upon publication to support full reproducibility.

\bibliographystyle{IEEEtran}
\bibliography{main}
\onecolumn
\section*{Supplementary Material}
\renewcommand{\thetable}{S\arabic{table}}
\renewcommand{\theHtable}{S\arabic{table}}
\begingroup
\fontsize{7}{8.2}\selectfont
\setlength{\tabcolsep}{3pt}
\renewcommand{\arraystretch}{1.05}
\setcounter{table}{0}
\begin{longtable}{lrrrrrrlll}
\caption{DB3 clinical metadata and membership in the fixed formal cohorts.}\label{tab:S1}\\
\toprule
Subject & Forearm (\%) & PLS & Cosmetic (y) & Kinematic (y) & Myoelectric (y) & EMG & B17 & BC40 & Formal subset\\\midrule
\endfirsthead
\multicolumn{10}{l}{Supplementary Table S1 continued}\\\toprule
Subject & Forearm (\%) & PLS & Cosmetic (y) & Kinematic (y) & Myoelectric (y) & EMG & B17 & BC40 & Formal subset\\\midrule
\endhead
\midrule
\endfoot
\bottomrule
\endlastfoot
s1 & 50 & 2 & 0 & 0 & 13 & 12 & Yes & No & Holdout3\\
s2 & 70 & 5 & 6 & 0 & 0 & 12 & Yes & Yes & Core6\\
s3 & 30 & 2 & 0 & 8 & 8 & 12 & Yes & No & Holdout3\\
s4 & 40 & 1 & 0 & 0 & 0 & 12 & Yes & Yes & Core6\\
s5 & 90 & 2 & 0 & 0.4 & 0 & 12 & Yes & Yes & Core6\\
s6 & 40 & 4 & 0 & 12 & 0 & 10 & No & No & Excluded: 10 effective EMG channels\\
s7 & 0 & 0 & 0 & 0 & 6 & 10 & No & No & Excluded: 10 effective EMG channels\\
s8 & 50 & 2 & 0 & 0 & 4 & 12 & Yes & Yes & Core6\\
s9 & 90 & 5 & 0 & 0 & 14 & 12 & Yes & Yes & Core6\\
s10 & 50 & 5 & 0 & 1.66 & 0 & 12 & Yes & No & Holdout3\\
s11 & 90 & 4 & 0 & 5 & 5 & 12 & Yes & Yes & Core6\\
\end{longtable}
PLS denotes phantom-limb sensation intensity. EMG counts nonconstant channels. Subjects s6 and s7 have identically zero EMG columns 9 and 10 and were excluded under the 12-effective-channel criterion. All9 comprises Core6 and Holdout3. BC40 was scheduled only for Core6; s1 lacks complete Exercise C recordings. Membership is the frozen experimental schedule, not a post-hoc selection. Metadata: \texttt{stage3/clinical\_metadata.json}; channel-ID audit: \texttt{reports/stage3/db3\_subject\_id\_discrepancy.json}.
\endgroup

\clearpage
\begingroup
\fontsize{7}{8}\selectfont
\setlength{\tabcolsep}{3pt}
\renewcommand{\arraystretch}{1.0}
\setcounter{table}{1}
Each cell shows accuracy on the upper line and Macro-F1 on the lower line, both in percent as participant mean $\pm$ participant SD after averaging three seeds and three corruption draws. A dash denotes an unevaluated condition. Subcohorts overlap their parent cohorts and are not independent replications. DB2 s6 has nine historical ACC36 runs; Holdout5 denotes the fixed zero-retuning confirmation split, not an entirely unseen participant history. Corresponding pairwise tests are in Table S3.
\medskip
\begin{longtable}{lccccccccc}
\caption{Complete condition-level classification results, including nested development and confirmation cohorts.}\label{tab:S2}\\
\toprule
Model & Clean & Burst-25 & Dropout-25 & Liftoff-25 & Burst-50 & Dropout-50 & Liftoff-50 & EMG-loss & ACC-loss\\\midrule
\endfirsthead
\multicolumn{10}{l}{Supplementary Table S2 continued}\\\toprule
Model & Clean & Burst-25 & Dropout-25 & Liftoff-25 & Burst-50 & Dropout-50 & Liftoff-50 & EMG-loss & ACC-loss\\\midrule
\endhead
\midrule
\endfoot
\bottomrule
\endlastfoot
\multicolumn{10}{l}{\textbf{DB2-All10, B17, n = 10}}\\\midrule
ACC-Only & \shortstack{65.19$\pm$6.51\\63.81$\pm$6.71} & \shortstack{65.19$\pm$6.51\\63.81$\pm$6.71} & \shortstack{65.19$\pm$6.51\\63.81$\pm$6.71} & \shortstack{65.19$\pm$6.51\\63.81$\pm$6.71} & \shortstack{65.19$\pm$6.51\\63.81$\pm$6.71} & \shortstack{65.19$\pm$6.51\\63.81$\pm$6.71} & \shortstack{65.19$\pm$6.51\\63.81$\pm$6.71} & \shortstack{65.19$\pm$6.51\\63.81$\pm$6.71} & \shortstack{6.80$\pm$1.76\\0.75$\pm$0.18}\\
Vanilla-Clean & \shortstack{80.92$\pm$5.83\\80.72$\pm$5.79} & \shortstack{50.57$\pm$7.62\\49.95$\pm$7.28} & \shortstack{70.38$\pm$6.70\\69.72$\pm$6.28} & \shortstack{30.91$\pm$4.57\\29.99$\pm$4.48} & \shortstack{43.91$\pm$6.30\\42.95$\pm$6.13} & \shortstack{58.41$\pm$7.33\\57.11$\pm$6.13} & \shortstack{29.93$\pm$4.54\\28.69$\pm$4.42} & \shortstack{28.76$\pm$10.31\\15.08$\pm$6.05} & \shortstack{30.74$\pm$10.69\\23.64$\pm$11.33}\\
Vanilla-Aug & \shortstack{82.63$\pm$6.55\\82.64$\pm$6.41} & \shortstack{78.49$\pm$7.71\\78.48$\pm$7.57} & \shortstack{77.93$\pm$6.74\\77.70$\pm$6.51} & \shortstack{67.58$\pm$7.18\\66.72$\pm$7.00} & \shortstack{76.67$\pm$8.11\\76.53$\pm$8.02} & \shortstack{71.03$\pm$7.05\\70.26$\pm$6.73} & \shortstack{66.15$\pm$6.87\\65.01$\pm$6.77} & \shortstack{64.56$\pm$6.94\\62.81$\pm$7.35} & \shortstack{59.83$\pm$9.45\\59.93$\pm$9.16}\\
Gate-Only & \shortstack{87.58$\pm$3.75\\87.43$\pm$3.69} & \shortstack{84.26$\pm$4.57\\84.04$\pm$4.50} & \shortstack{84.05$\pm$4.54\\83.78$\pm$4.48} & \shortstack{84.25$\pm$4.56\\84.01$\pm$4.44} & \shortstack{79.61$\pm$5.20\\79.20$\pm$5.06} & \shortstack{79.28$\pm$5.31\\78.85$\pm$5.19} & \shortstack{79.57$\pm$5.33\\79.15$\pm$5.14} & \shortstack{63.62$\pm$6.94\\61.61$\pm$7.41} & \shortstack{66.68$\pm$6.59\\66.40$\pm$6.49}\\
\end{longtable}
\endgroup

\clearpage
\begingroup
\fontsize{7}{8.2}\selectfont
\setlength{\tabcolsep}{3pt}
\renewcommand{\arraystretch}{1.05}
\setcounter{table}{2}
\begin{longtable}{llrrcrcrrrrr}
\caption{Independent recomputation of participant-paired QA-CMIF-minus-baseline contrasts.}\label{tab:S3}\\
\toprule
Condition & Baseline & $\Delta$ & SD & 95\% CI & $d_z$ & W/L/T & $p_t$ raw & $p_W$ raw & $p_t$ primary & $p_W$ primary & $p_t$ full\\\midrule
\endfirsthead
\multicolumn{12}{l}{Supplementary Table S3 continued}\\\toprule
Condition & Baseline & $\Delta$ & SD & 95\% CI & $d_z$ & W/L/T & $p_t$ raw & $p_W$ raw & $p_t$ primary & $p_W$ primary & $p_t$ full\\\midrule
\endhead
\midrule
\endfoot
\bottomrule
\endlastfoot
\multicolumn{12}{l}{\textbf{DB2-All10, B17, Accuracy}}\\\midrule
Clean & ACC-Only & 22.53 & 5.31 & [18.73, 26.33] & 4.24 & 10/0/0 & $2.95\times10^{-7}$ & 0.0020 & -- & -- & --\\
Clean & Vanilla-Clean & 6.80 & 3.30 & [4.44, 9.16] & 2.06 & 10/0/0 & $1.10\times10^{-4}$ & 0.0020 & -- & -- & 0.0054\\
Clean & Vanilla-Aug & 5.09 & 3.58 & [2.53, 7.65] & 1.42 & 10/0/0 & 0.0015 & 0.0020 & -- & -- & 0.0624\\
Clean & Gate-Only & 0.13 & 0.48 & [-0.21, 0.47] & 0.28 & 6/4/0 & 0.3987 & 0.4316 & -- & -- & 1.0000\\
Clean & w/o Bypass & -1.13 & 1.39 & [-2.12, -0.13] & -0.81 & 3/7/0 & 0.0303 & 0.0371 & -- & -- & 0.9685\\
Clean & w/o Stop-Grad & 0.08 & 0.22 & [-0.08, 0.23] & 0.36 & 6/3/1 & 0.2904 & 0.2031 & -- & -- & 1.0000\\
Clean & w/o Align Loss & -0.02 & 0.40 & [-0.31, 0.27] & -0.05 & 5/5/0 & 0.8860 & 0.7695 & -- & -- & 1.0000\\
\end{longtable}
All differences are percentage points. Tests are two-sided. W/L/T denotes positive, negative, and tied participant differences. The primary Holm family comprises the three 50\%-corruption modes against five non-ablation baselines, or three baselines on ABC52: 15 or 9 contrasts. Full-table Holm includes all non-ACC-only accuracy contrasts within the same cohort and task; separate Macro-F1 families are computed independently and are descriptive additions. Values outside a primary family have no primary-adjusted entry. Exact Wilcoxon values enumerate signed ranks and discard zero differences. Machine-readable Table S3 and full Wilcoxon adjustments are supplied in the accompanying ancillary CSV.
\endgroup

\clearpage\setcounter{table}{3}
\begin{table*}[!htbp]
\centering\begingroup
\fontsize{7.5}{8.5}\selectfont
\setlength{\tabcolsep}{3pt}
\renewcommand{\arraystretch}{0.95}
\caption{Full ablation results, including w/o Stop-Grad (B17).}
\label{tab:S4}
\setlength{\TableUsable}{\dimexpr\textwidth-54pt\relax}
\begin{tabular}{@{}>{\centering\arraybackslash}p{0.14\TableUsable}>{\centering\arraybackslash}p{0.085\TableUsable}>{\centering\arraybackslash}p{0.085\TableUsable}>{\centering\arraybackslash}p{0.085\TableUsable}>{\centering\arraybackslash}p{0.085\TableUsable}>{\centering\arraybackslash}p{0.12\TableUsable}>{\centering\arraybackslash}p{0.06\TableUsable}>{\centering\arraybackslash}p{0.12\TableUsable}>{\centering\arraybackslash}p{0.12\TableUsable}>{\centering\arraybackslash}p{0.1\TableUsable}@{}}
\toprule
Model & clean Acc & Burst-50\% & Dropout-50\% & Liftoff-50\% & \shortstack{Gate\\AUROC/F1\\Burst-50\%} & \shortstack{Clean\\FAR (\%)} & \shortstack{Clean dispersion\\$\sigma^2_{\rm clean}$} & \shortstack{MSE reduction\\Liftoff-50\%\\(\%)} & Learned $\gamma$\\\midrule
\midrule\multicolumn{10}{@{}l}{\strong{DB3-All9}  ($n = 9$ amputees,   $2000\text{ Hz}$)}\\
Vanilla-Aug & \shortstack{68.13\\$\pm$ 12.81} & \shortstack{63.70\\$\pm$ 14.53} & \shortstack{60.49\\$\pm$ 13.99} & \shortstack{59.33\\$\pm$ 15.16} & — & — & — & — & —\\
Gate-Only & \shortstack{74.93\\$\pm$ 12.39} & \shortstack{68.67\\$\pm$ 13.81} & \shortstack{68.65\\$\pm$ 13.77} & \shortstack{68.48\\$\pm$ 13.82} & \shortstack{0.9991\\ / 0.9904} & 0.41 & \shortstack{1.405\\$\pm$ 0.194} & \shortstack{0.00\\$\pm$ 0.00} & —\\
w/o Bypass & \shortstack{75.23\\$\pm$ 11.51} & \shortstack{69.03\\$\pm$ 13.34} & \shortstack{68.98\\$\pm$ 13.55} & \shortstack{69.00\\$\pm$ 13.44} & \shortstack{0.9989\\ / 0.9863} & 0.35 & \shortstack{0.384\\$\pm$ 0.038} & \shortstack{+37.09\\$\pm$ 11.28} & 0.500 (fixed)\\
w/o Align Loss & \shortstack{74.93\\$\pm$ 12.31} & \shortstack{69.46\\$\pm$ 13.41} & \shortstack{69.50\\$\pm$ 13.46} & \shortstack{69.39\\$\pm$ 13.48} & \shortstack{0.9992\\ / 0.9908} & 0.41 & \shortstack{1.364\\$\pm$ 0.185} & \shortstack{-9.79\\$\pm$ 6.70} & \shortstack{0.655\\$\pm$ 0.050}\\
w/o Stop-Grad ($\alpha$ not detached) & \shortstack{74.86\\$\pm$ 12.40} & \shortstack{70.37\\$\pm$ 13.42} & \shortstack{69.94\\$\pm$ 13.42} & \shortstack{70.11\\$\pm$ 13.51} & \shortstack{0.9990\\ / 0.9870} & 0.58 & \shortstack{1.277\\$\pm$ 0.193} & \shortstack{+43.41\\$\pm$ 6.82} & \shortstack{0.646\\$\pm$ 0.040}\\
\textbf{QA-CMIF} & \shortstack{74.85\\$\pm$ 12.63} & \shortstack{69.97\\$\pm$ 13.48} & \shortstack{69.97\\$\pm$ 13.48} & \shortstack{70.04\\$\pm$ 13.49} & \shortstack{0.9992\\ / 0.9898} & 0.43 & \shortstack{1.286\\$\pm$ 0.189} & \shortstack{+42.43\\$\pm$ 6.90} & \shortstack{0.625\\$\pm$ 0.037}\\
\midrule\multicolumn{10}{@{}l}{\strong{DB2-All10}  ($n = 10$ intact,   $2000\text{ Hz}$)}\\
Vanilla-Aug & \shortstack{82.63\\$\pm$ 6.55} & \shortstack{76.67\\$\pm$ 8.11} & \shortstack{71.03\\$\pm$ 7.05} & \shortstack{66.15\\$\pm$ 6.87} & — & — & — & — & —\\
Gate-Only & \shortstack{87.58\\$\pm$ 3.75} & \shortstack{79.61\\$\pm$ 5.20} & \shortstack{79.28\\$\pm$ 5.31} & \shortstack{79.57\\$\pm$ 5.33} & \shortstack{0.9991\\ / 0.9897} & 0.41 & \shortstack{1.350\\$\pm$ 0.131} & \shortstack{0.00\\$\pm$ 0.00} & —\\
w/o Bypass & \shortstack{88.84\\$\pm$ 3.47} & \shortstack{80.61\\$\pm$ 5.06} & \shortstack{80.27\\$\pm$ 4.84} & \shortstack{80.37\\$\pm$ 4.94} & \shortstack{0.9979\\ / 0.9733} & 0.44 & \shortstack{0.358\\$\pm$ 0.039} & \shortstack{+36.89\\$\pm$ 10.30} & 0.500 (fixed)\\
w/o Align Loss & \shortstack{87.74\\$\pm$ 3.50} & \shortstack{80.75\\$\pm$ 4.97} & \shortstack{80.71\\$\pm$ 5.09} & \shortstack{80.91\\$\pm$ 5.03} & \shortstack{0.9992\\ / 0.9904} & 0.42 & \shortstack{1.306\\$\pm$ 0.128} & \shortstack{-7.78\\$\pm$ 3.03} & \shortstack{0.628\\$\pm$ 0.024}\\
w/o Stop-Grad ($\alpha$ not detached) & \shortstack{87.64\\$\pm$ 3.57} & \shortstack{81.83\\$\pm$ 4.83} & \shortstack{80.89\\$\pm$ 4.74} & \shortstack{81.36\\$\pm$ 4.98} & \shortstack{0.9979\\ / 0.9770} & 0.72 & \shortstack{1.212\\$\pm$ 0.125} & \shortstack{+41.61\\$\pm$ 5.39} & \shortstack{0.654\\$\pm$ 0.062}\\
\textbf{QA-CMIF} & \shortstack{87.72\\$\pm$ 3.66} & \shortstack{81.28\\$\pm$ 4.72} & \shortstack{81.03\\$\pm$ 4.78} & \shortstack{81.26\\$\pm$ 4.86} & \shortstack{0.9989\\ / 0.9868} & 0.42 & \shortstack{1.242\\$\pm$ 0.118} & \shortstack{+39.84\\$\pm$ 5.20} & \shortstack{0.595\\$\pm$ 0.032}\\
\midrule\multicolumn{10}{@{}l}{\strong{DB5-All10}  ($n = 10$ intact,   $200\text{ Hz}$)}\\
Vanilla-Aug & \shortstack{80.49\\$\pm$ 4.30} & \shortstack{72.26\\$\pm$ 5.74} & \shortstack{68.71\\$\pm$ 4.67} & \shortstack{63.63\\$\pm$ 6.96} & — & — & — & — & —\\
Gate-Only & \shortstack{83.04\\$\pm$ 4.18} & \shortstack{75.78\\$\pm$ 4.97} & \shortstack{75.50\\$\pm$ 4.86} & \shortstack{75.46\\$\pm$ 4.83} & \shortstack{0.9959\\ / 0.9725} & 0.42 & \shortstack{1.237\\$\pm$ 0.034} & \shortstack{0.00\\$\pm$ 0.00} & —\\
w/o Bypass & \shortstack{84.11\\$\pm$ 4.58} & \shortstack{76.59\\$\pm$ 4.76} & \shortstack{76.77\\$\pm$ 4.71} & \shortstack{76.84\\$\pm$ 4.60} & \shortstack{0.9910\\ / 0.9429} & 0.51 & \shortstack{0.315\\$\pm$ 0.008} & \shortstack{+43.37\\$\pm$ 4.94} & 0.100 (fixed)\\
w/o Align Loss & \shortstack{83.14\\$\pm$ 4.12} & \shortstack{77.08\\$\pm$ 4.59} & \shortstack{76.98\\$\pm$ 4.46} & \shortstack{77.03\\$\pm$ 4.46} & \shortstack{0.9960\\ / 0.9737} & 0.43 & \shortstack{1.200\\$\pm$ 0.032} & \shortstack{+3.91\\$\pm$ 1.02} & \shortstack{0.448\\$\pm$ 0.019}\\
w/o Stop-Grad ($\alpha$ not detached) & \shortstack{83.05\\$\pm$ 4.02} & \shortstack{78.28\\$\pm$ 4.42} & \shortstack{78.02\\$\pm$ 4.35} & \shortstack{78.16\\$\pm$ 4.36} & \shortstack{0.9954\\ / 0.9698} & 0.47 & \shortstack{1.140\\$\pm$ 0.040} & \shortstack{+42.26\\$\pm$ 2.10} & \shortstack{0.453\\$\pm$ 0.018}\\
\textbf{QA-CMIF} & \shortstack{83.08\\$\pm$ 4.02} & \shortstack{78.18\\$\pm$ 4.42} & \shortstack{78.01\\$\pm$ 4.34} & \shortstack{78.11\\$\pm$ 4.37} & \shortstack{0.9953\\ / 0.9694} & 0.43 & \shortstack{1.140\\$\pm$ 0.041} & \shortstack{+41.59\\$\pm$ 2.05} & \shortstack{0.451\\$\pm$ 0.018}\\
\bottomrule\end{tabular}
\par\vspace{2pt}
\begin{minipage}{\textwidth}\scriptsize
$^{\text{a}}$ Under Dropout-50\% and Liftoff-50\%, QA-CMIF achieves Gate AUROC $\ge 0.9995$ and Gate F1 $\ge 0.9960$ across all three cohorts. Figs. \ref{fig:4} and \ref{fig:5} illustrate the paired clean representation variance $\sigma_{\text{clean}}^2$ and bad-channel MSE reduction $\text{MSE}_{\text{red}}(\%)$ across ablation variants.\par Accuracy values are percentages. Diagnostic values are descriptive; no diagnostic entries are bolded. 
\end{minipage}\endgroup
\end{table*}
\par\medskip
\begingroup\small
\noindent\textbf{Supplementary Table S4 continued: DB5-All10 ABC52 latent-repair diagnostics.}
\begin{center}\begin{tabular}{llcc}\toprule
Model & Condition & Accuracy (\%) & Bad-channel MSE reduction (\%)\\\midrule
Gate-Only & Liftoff-50 & 69.68$\pm$4.86 & 0.00$\pm$0.00\\
Gate-Only & Burst-50 & 69.68$\pm$4.75 & 0.00$\pm$0.00\\
Gate-Only & Dropout-50 & 69.49$\pm$4.92 & 0.00$\pm$0.00\\
QA-CMIF & Liftoff-50 & 73.76$\pm$4.61 & 50.73$\pm$4.05\\
QA-CMIF & Burst-50 & 73.33$\pm$4.54 & 48.53$\pm$4.51\\
QA-CMIF & Dropout-50 & 73.64$\pm$4.61 & 50.80$\pm$4.12\\
\bottomrule\end{tabular}\end{center}
Values are participant mean $\pm$ SD over ten participants after three-seed averaging. These rows document the ABC52 diagnostic range cited in Section V-D; no ABC52 ablation training is implied.
\endgroup

\clearpage\begin{table*}[!htbp]
\centering
\begingroup
\fontsize{8}{9.5}\selectfont
\setlength{\tabcolsep}{2pt}
\renewcommand{\arraystretch}{1.12}
\setlength{\TableUsable}{\dimexpr\textwidth-32pt\relax}
\caption{EXPLORATORY PAIRED TASK-COMPLEXITY EXPANSION ANALYSIS OF QA-CMIF VS. GATE-ONLY (BC40 VS. B17 ON DB3 AMPUTEES AND ABC52 VS. B17 ON DB5 INTACT SUBJECTS)}
\label{tab:S5}
\begin{tabular}{@{}>{\raggedright\arraybackslash}p{0.130000\TableUsable}>{\raggedright\arraybackslash}p{0.100000\TableUsable}>{\raggedright\arraybackslash}p{0.120000\TableUsable}>{\raggedright\arraybackslash}p{0.120000\TableUsable}>{\raggedright\arraybackslash}p{0.130000\TableUsable}>{\raggedright\arraybackslash}p{0.120000\TableUsable}>{\raggedright\arraybackslash}p{0.070000\TableUsable}>{\raggedright\arraybackslash}p{0.100000\TableUsable}>{\raggedright\arraybackslash}p{0.110000\TableUsable}@{}}
\toprule
\TCell{V}{0}{0}{Cohort \& Expansion Pair} & \TCell{V}{0}{1}{Condition} & \TCell{V}{0}{2}{B17 (17 cl.) Gain\newline QA-CMIF \ensuremath{-} Gate-Only ($\text{pp}$)} & \TCell{V}{0}{3}{40/52-Class Gain\newline QA-CMIF \ensuremath{-} Gate-Only ($\text{pp}$)} & \TCell{V}{0}{4}{Paired Expansion\newline  $\Delta_{\text{expand}} \pm \text{SD}$ ($\text{pp}$)} & \TCell{V}{0}{5}{95\% CI ($\text{pp}$)} & \TCell{V}{0}{6}{Positive Subjects} & \TCell{V}{0}{7}{Paired $t$ ($d_z$)} & \TCell{V}{0}{8}{Raw $p$-value (Wilcoxon $p$)} \\
\midrule
\TCell{V}{1}{0}{\strong{DB3-Core6}\newline ($n = 6$ amputees)\newline BC40 (40 cl.) vs. B17 (17 cl.)} & \TCell{V}{1}{1}{clean} & \TCell{V}{1}{2}{+0.01 $\pm$ 0.46} & \TCell{V}{1}{3}{+0.36 $\pm$ 0.22} & \TCell{V}{1}{4}{+0.35 $\pm$ 0.46} & \TCell{V}{1}{5}{$[-0.13, +0.84]$} & \TCell{V}{1}{6}{4/6} & \TCell{V}{1}{7}{+1.873 (+0.765)} & \TCell{V}{1}{8}{0.120 (0.156)} \\
\TCell{V}{2}{0}{} & \TCell{V}{2}{1}{Burst-25\%} & \TCell{V}{2}{2}{+0.49 $\pm$ 0.56} & \TCell{V}{2}{3}{+1.55 $\pm$ 0.33} & \TCell{V}{2}{4}{\strong{+1.06 $\pm$ 0.47}} & \TCell{V}{2}{5}{$[+0.57, +1.55]$} & \TCell{V}{2}{6}{\strong{6/6}} & \TCell{V}{2}{7}{+5.529 (+2.257)} & \TCell{V}{2}{8}{\strong{0.00265 (0.0313)}} \\
\TCell{V}{3}{0}{} & \TCell{V}{3}{1}{Dropout-25\%} & \TCell{V}{3}{2}{+0.44 $\pm$ 0.56} & \TCell{V}{3}{3}{+1.58 $\pm$ 0.27} & \TCell{V}{3}{4}{\strong{+1.14 $\pm$ 0.48}} & \TCell{V}{3}{5}{$[+0.63, +1.65]$} & \TCell{V}{3}{6}{\strong{6/6}} & \TCell{V}{3}{7}{+5.784 (+2.361)} & \TCell{V}{3}{8}{\strong{0.00217 (0.0313)}} \\
\TCell{V}{4}{0}{} & \TCell{V}{4}{1}{Liftoff-25\%} & \TCell{V}{4}{2}{+0.55 $\pm$ 0.57} & \TCell{V}{4}{3}{+1.58 $\pm$ 0.31} & \TCell{V}{4}{4}{\strong{+1.03 $\pm$ 0.49}} & \TCell{V}{4}{5}{$[+0.52, +1.55]$} & \TCell{V}{4}{6}{\strong{6/6}} & \TCell{V}{4}{7}{+5.152 (+2.103)} & \TCell{V}{4}{8}{\strong{0.00361 (0.0313)}} \\
\TCell{V}{5}{0}{} & \TCell{V}{5}{1}{Burst-50\%} & \TCell{V}{5}{2}{+1.17 $\pm$ 0.62} & \TCell{V}{5}{3}{+2.59 $\pm$ 0.36} & \TCell{V}{5}{4}{\strong{+1.43 $\pm$ 0.62}} & \TCell{V}{5}{5}{$[+0.78, +2.07]$} & \TCell{V}{5}{6}{\strong{6/6}} & \TCell{V}{5}{7}{+5.672 (+2.316)} & \TCell{V}{5}{8}{\strong{0.00237 (0.0313)}} \\
\TCell{V}{6}{0}{} & \TCell{V}{6}{1}{Dropout-50\%} & \TCell{V}{6}{2}{+1.22 $\pm$ 0.69} & \TCell{V}{6}{3}{+2.60 $\pm$ 0.14} & \TCell{V}{6}{4}{\strong{+1.37 $\pm$ 0.57}} & \TCell{V}{6}{5}{$[+0.77, +1.98]$} & \TCell{V}{6}{6}{\strong{6/6}} & \TCell{V}{6}{7}{+5.868 (+2.395)} & \TCell{V}{6}{8}{\strong{0.00204 (0.0313)}} \\
\TCell{V}{7}{0}{} & \TCell{V}{7}{1}{Liftoff-50\%} & \TCell{V}{7}{2}{+1.45 $\pm$ 0.61} & \TCell{V}{7}{3}{+2.69 $\pm$ 0.25} & \TCell{V}{7}{4}{\strong{+1.24 $\pm$ 0.50}} & \TCell{V}{7}{5}{$[+0.71, +1.77]$} & \TCell{V}{7}{6}{\strong{6/6}} & \TCell{V}{7}{7}{+6.001 (+2.450)} & \TCell{V}{7}{8}{\strong{0.00185 (0.0313)}} \\
\midrule
\TCell{V}{8}{0}{\strong{DB5-All10}\newline ($n = 10$ intact)\newline ABC52 (52 cl.) vs. B17 (17 cl.)} & \TCell{V}{8}{1}{clean} & \TCell{V}{8}{2}{+0.05 $\pm$ 0.33} & \TCell{V}{8}{3}{+0.26 $\pm$ 0.20} & \TCell{V}{8}{4}{+0.21 $\pm$ 0.37} & \TCell{V}{8}{5}{$[-0.05, +0.48]$} & \TCell{V}{8}{6}{8/10} & \TCell{V}{8}{7}{+1.830 (+0.579)} & \TCell{V}{8}{8}{0.101 (0.105)} \\
\TCell{V}{9}{0}{} & \TCell{V}{9}{1}{Burst-25\%} & \TCell{V}{9}{2}{+1.03 $\pm$ 0.56} & \TCell{V}{9}{3}{+2.05 $\pm$ 0.51} & \TCell{V}{9}{4}{\strong{+1.01 $\pm$ 0.55}} & \TCell{V}{9}{5}{$[+0.62, +1.41]$} & \TCell{V}{9}{6}{\strong{9/10}} & \TCell{V}{9}{7}{+5.803 (+1.835)} & \TCell{V}{9}{8}{\strong{$2.59 \times 10^{-4}$ (0.00391)}} \\
\TCell{V}{10}{0}{} & \TCell{V}{10}{1}{Liftoff-25\%} & \TCell{V}{10}{2}{+1.29 $\pm$ 0.61} & \TCell{V}{10}{3}{+2.32 $\pm$ 0.63} & \TCell{V}{10}{4}{\strong{+1.03 $\pm$ 0.49}} & \TCell{V}{10}{5}{$[+0.67, +1.38]$} & \TCell{V}{10}{6}{\strong{10/10}} & \TCell{V}{10}{7}{+6.582 (+2.081)} & \TCell{V}{10}{8}{\strong{$1.01 \times 10^{-4}$ (0.00195)}} \\
\TCell{V}{11}{0}{} & \TCell{V}{11}{1}{Burst-50\%} & \TCell{V}{11}{2}{+2.40 $\pm$ 0.86} & \TCell{V}{11}{3}{+3.65 $\pm$ 0.75} & \TCell{V}{11}{4}{\strong{+1.25 $\pm$ 0.60}} & \TCell{V}{11}{5}{$[+0.82, +1.69]$} & \TCell{V}{11}{6}{\strong{9/10}} & \TCell{V}{11}{7}{+6.564 (+2.076)} & \TCell{V}{11}{8}{\strong{$1.03 \times 10^{-4}$ (0.00391)}} \\
\TCell{V}{12}{0}{} & \TCell{V}{12}{1}{Dropout-50\%} & \TCell{V}{12}{2}{+2.52 $\pm$ 0.93} & \TCell{V}{12}{3}{+4.15 $\pm$ 0.95} & \TCell{V}{12}{4}{\strong{+1.63 $\pm$ 0.69}} & \TCell{V}{12}{5}{$[+1.14, +2.13]$} & \TCell{V}{12}{6}{\strong{10/10}} & \TCell{V}{12}{7}{+7.437 (+2.352)} & \TCell{V}{12}{8}{\strong{$3.95 \times 10^{-5}$ (0.00195)}} \\
\TCell{V}{13}{0}{} & \TCell{V}{13}{1}{Liftoff-50\%} & \TCell{V}{13}{2}{+2.65 $\pm$ 0.77} & \TCell{V}{13}{3}{+4.08 $\pm$ 0.98} & \TCell{V}{13}{4}{\strong{+1.43 $\pm$ 0.70}} & \TCell{V}{13}{5}{$[+0.93, +1.93]$} & \TCell{V}{13}{6}{\strong{9/10}} & \TCell{V}{13}{7}{+6.469 (+2.046)} & \TCell{V}{13}{8}{\strong{$1.15 \times 10^{-4}$ (0.00391)}} \\
\bottomrule
\end{tabular}
\endgroup
\end{table*}

\end{document}